\documentclass[10pt,twocolumn,letterpaper]{article}

\usepackage{cvpr}
\usepackage{times}
\usepackage{epsfig}
\usepackage{graphicx}
\usepackage{amsmath}
\usepackage{amssymb}
\usepackage{bm}
\usepackage{chngpage}

\usepackage[breaklinks=true,bookmarks=false]{hyperref}

\newcommand{\PutCapDataset}{\put(2,3){{~~~~~~~~~~~~~~~~~~~~~~~~~~~~~~~~~~~~~~~OTB-2013 Dataset ~~~~~~~~~~~~~~~~~~~~~~~~~~~~~~~~~~~~~~~~~~~~~~~~~~~~~~~~~~~~~~~~~~~~~~ OTB-2015 Dataset }}}
\newcommand{\PutCapablation}{\put(2,3){{~~~~~~~~~~~~~~~~~~~~~~~~~~~~~~~~~~~~~~~~~~~~~~~~~~OTB-2015 Dataset~~~~~~~~~~~~~~~~~~~~~~~~~~~~~~~~~~~~~~~~~~~~~~~~~~~~~~~~~~~~~VOT-2016 Dataset}}}
\newcommand{\PutCapCombine}{\put(2,3){{~~~~~~~${\bf P}_d^1$~~~~~~~~~~~~~~~~~~~~${\bf P}_d^m$~~~~~~~~~~~~~~~~~~${\bf P}_d^9$~~~~~~~~~~${\bf v}_d$}}}

\cvprfinalcopy 

\def\cvprPaperID{****} 

\ifcvprfinal\fi
\begin{document}

\title{Correlation Tracking via Joint Discrimination and Reliability Learning}

\author{Chong Sun$^{1}$, Dong Wang$^1$\thanks{Corresponding Author}, Huchuan Lu$^1$, Ming-Hsuan Yang$^2$\\
$^1$School of Information and Communication Engineering, Dalian University of Technology, China\\ $^2$Electrical Engineering and Computer Science, University of California, Merced, USA\\
{\tt\small waynecool@mail.dlut.edu.cn, \{wdice,lhchuan\}@dlut.edu.cn,  mhyang@ucmerced.edu}
}

\maketitle
\thispagestyle{empty}

\begin{abstract}
For visual tracking, an ideal filter learned by the correlation filter (CF) method should take both discrimination and reliability information. However, existing attempts usually focus on the former one while pay less attention to reliability learning. This may make the learned filter be dominated by the unexpected salient regions on the feature map, thereby resulting in model degradation. To address this issue, we propose a novel CF-based optimization problem to jointly model the discrimination and reliability information. First, we treat the filter as the element-wise product of a base filter and a reliability term. The base filter is aimed to learn the discrimination information between the target and backgrounds, and the reliability term encourages the final filter to focus on more reliable regions. Second, we introduce a local response consistency regular term to emphasize equal contributions of different regions and avoid the tracker being dominated by unreliable regions. The proposed optimization problem can be solved using the alternating direction method and speeded up in the Fourier domain. We conduct extensive experiments on the OTB-2013, OTB-2015 and VOT-2016 datasets to evaluate the proposed tracker. Experimental results show that our tracker performs favorably against other state-of-the-art trackers.

\end{abstract}

\begin{figure}[t]
\centering
\begin{tabular}{c@{}c}

\includegraphics[width=0.47\linewidth,height=14.5mm]{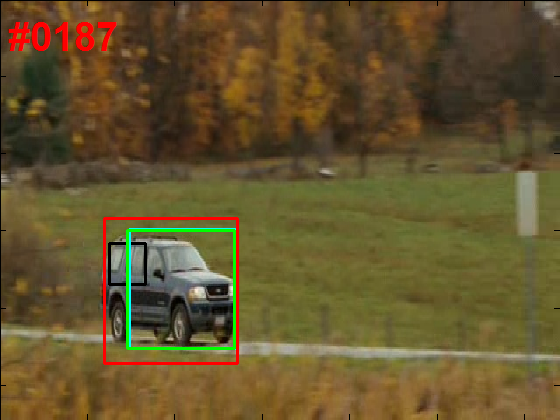}
\ &
\includegraphics[width=0.47\linewidth,height=14.5mm]{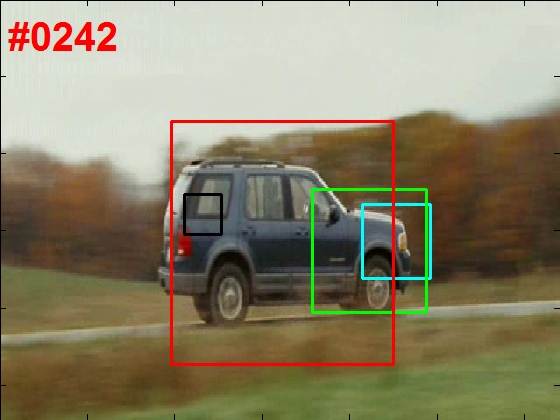}
\ \\
\includegraphics[width=0.47\linewidth,height=14.5mm]{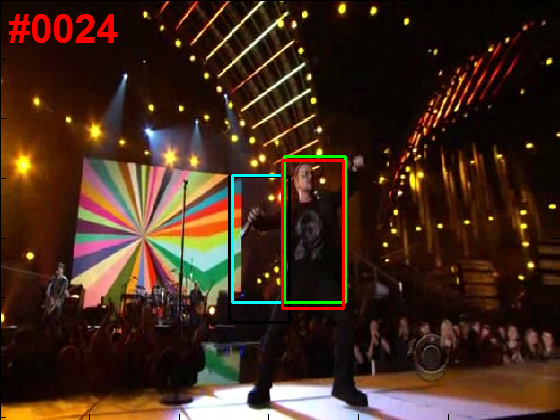}
\ &
\includegraphics[width=0.47\linewidth,height=14.5mm]{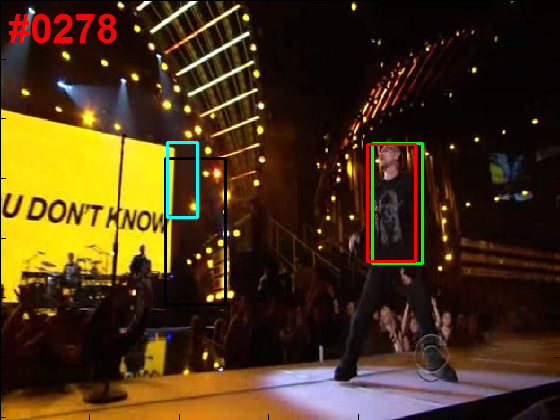}
\ \\
\includegraphics[width=0.47\linewidth,height=14.5mm]{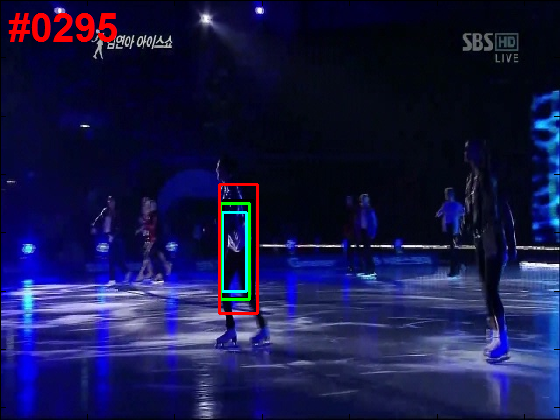}
\ &
\includegraphics[width=0.47\linewidth,height=14.5mm]{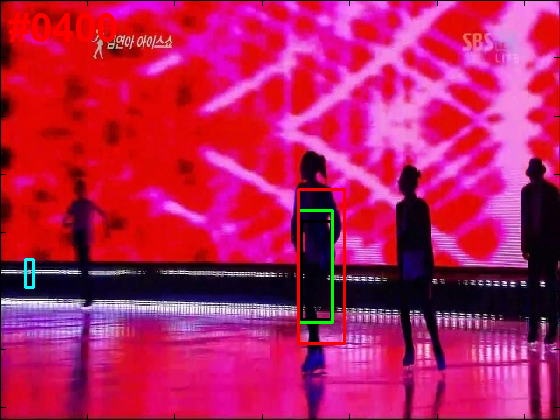}
\ \\
\end{tabular}
\begin{tabular}{c}
{\kern 6mm}\includegraphics[width=0.7\linewidth]{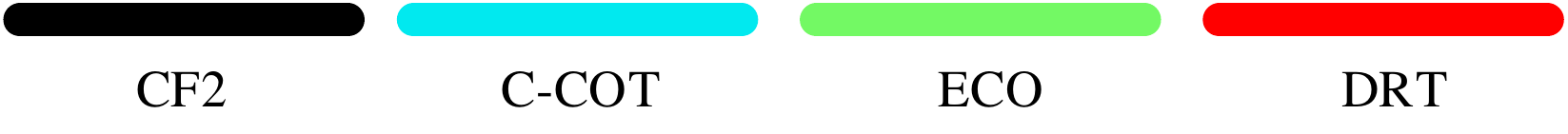}
\end{tabular}

\caption{Example tracking results of different methods on the OTB dataset. Our tracker (DRT)
has comparable or better results compared with the existing best tracker ECO.}
\label{fig:tracking_results}
\vspace{-2mm}
\end{figure}

\section{Introduction}
Visual tracking is a hot topic for its wide applications in many computer vision tasks, such as
video surveillance, behaviour analysis, augmented reality, to name a few.  Though many trackers~\cite{henriques2015high,qi2016hedged,lukezic2017discriminative,salsc2018,LiWWL18} have been
proposed to address this task, designing a robust visual tracking system is still challenging as the tracked
target may undergo large deformations, rotations and other challenges.

Numerous recent studies apply the correlation filter (CF) for robust visual tracking. With low computational load, the CF-based tracker can exploit
large numbers of cyclically shifted samples for learning, thus showing superior performance. However, as the correlation filter takes the entire image as the positive sample
and the cyclically shifted images as negative ones, the learned filters are likely to be influenced by the background regions. Existing methods
(\eg~\cite{danelljan2015learning,danelljan2016beyond,danelljan2016eco}) address this problem by incorporating a spatial regularization on the
filter, so that the learned filter weights focus on the central part of the target object. In~\cite{kiani2017learning}, the authors prove that the correlation filter method
can be used to simulate the conventional ridge regression method. By multiplying the filter with a binary mask, the tracker is able to generate the real training samples
having the same size as the target object, and thus better suppressing the background regions. However, this method has two limitations:
first, it exploits the augmented Lagrangian method for model learning,
which limits the model extension; second, even though the background region outside the bounding box is suppressed, the tracker may also be
influenced by the background region inside the bounding box.

\begin{figure}[t]
  \centering
  \begin{tabular}{c@{}c}
  \includegraphics[width=0.9\linewidth,height=25mm]{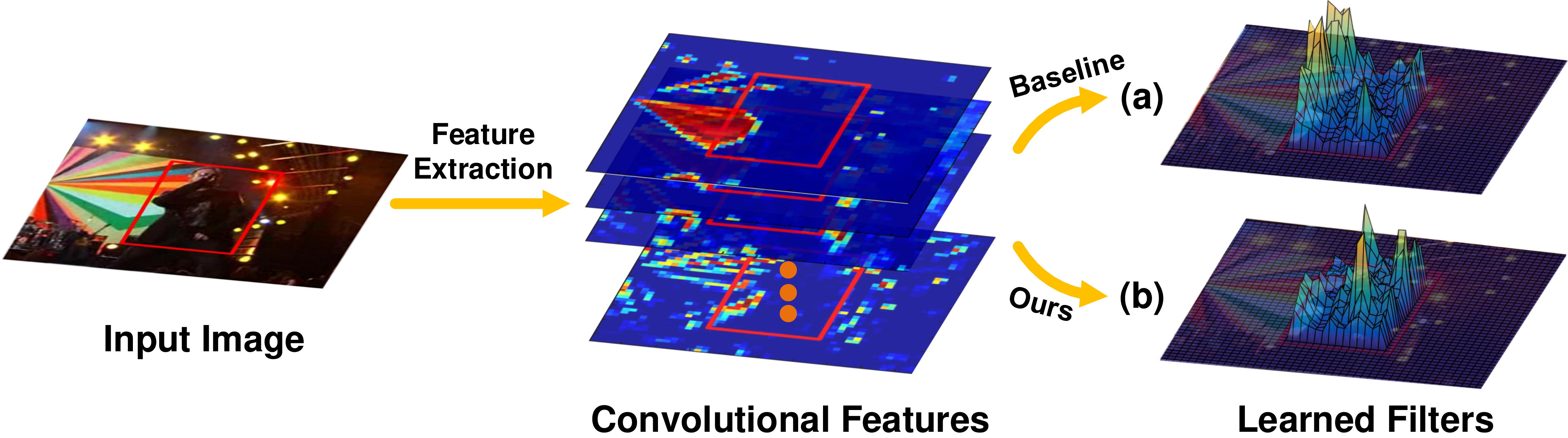}\\
  \end{tabular}
   \vspace{-1mm}
  \caption{Example showing that our learned filter coefficients are insusceptible to the response distribution of the feature map.
  In (a) and (b), we compute the square sum of filter coefficients
  across the channel dimension, and obtain a spatial energy distribution for the learned filter. (a) The baseline method,
  which does not consider the local consistency regular term and set $\beta_m, m=\{1,...,M\}$ as 1.
 (b) The proposed joint learning formula. Compared to our method, the baseline method learns large coefficients on the background
 (\ie the left-side region in the bounding box).}
 \label{fig:laplace_compare}
 \vspace{-3.0mm}
\end{figure}

With the great success of deep convolutional neural network (CNN) in object detection and classification,
more and more CF based trackers resort to the pre-trained CNN model
for robust target representation~\cite{wang2016stct,wang2015visual,danelljan2016eco}.
Since most of CNN models are pre-trained with respect to the task of object classification or detection,
they tend to retain the features useful for distinguishing different
categories of objects, and lose much information for instance level classification.
Thus, the responses of the feature map are usually sparsely and non-uniformly distributed, which
makes the learned filter weights inevitably highlight the high response regions.
In this case, the tracking results are dominated by such high response regions, while these regions are in fact
not always reliable (see Figure~\ref{fig:laplace_compare} for an example).

In this paper, we present a novel CF-based optimization problem to clearly learn the discrimination
and reliability information, and then develop an effective tracking method (denoted as DRT).
The concept of the base filter is proposed to focus on discrimination learning. To do this,
we introduce the local response consistency constraint into the traditional CF framework.
This constraint ensures that the responses generated by different sub-regions of the
base filter have small difference, thereby emphasizing the similar importance of each sub-region.
The reliability weight map is also considered in our formula. It is online jointly learned with the base
filter and is aimed at learning the reliability information.
The base filter and reliability term are jointly optimized by the alternating direction method, and
their element-wise product produces effective filter weights for the tracking task.
Finally, we conduct extensive experiments on three benchmark datasets to demonstrate the effectiveness
of our method  (see Figure~\ref{fig:tracking_results} and Section~\ref{sec:exp}).

Our contributions are four folds:

\begin{itemize}
\vspace{-2mm}
\item Our work is the first attempt to jointly model both discrimination and reliability information using
the correlation filter framework. We treat an ideal filter as the element-wise product of a base filter and
a reliability term and propose a novel optimization problem with insightful constraints.
\vspace{-2mm}
\item The local response consistency constraint is introduced to ensure that different sub-regions of the
 base filter have similar importance. Thus, the base filter will highlight the entire target even though the
 feature maps may be dominated by some specific regions.
\vspace{-2mm}
\item The reliability weight map is exploited to depict the importance of each sub-region in the filter
(i.e. reliability learning) and is online jointly learned with the base filter. Being insusceptible to the response
distributions of the feature map, it can better reflect the real tracking performance for different sub-regions.
\vspace{-2mm}
\item Our tracker achieves remarkable tracking performance on the OTB-2013, OTB-2015 and VOT-2016 benchmarks.
Our tracker has the best results on all the reported datasets.
\end{itemize}

\section{Relate Work}
Correlation filters (CF) have shown great success in visual tracking for their efficient learning process.
In this section, we briefly introduce the CF-based trackers that are closely related to our work.

The early CF-based trackers exploit a single feature channel as input, and thus usually
have very impressive tracking speed. The MOSSE tracker~\cite{bolme2010visual} exploits
the adaptive correlation filter, which optimizes the
sum of squared error.
Henriques~\etal~\cite{henriques2012exploiting} introduce the kernel trick into the correlation filter formula.
By exploiting the property of the circulant matrix,
they provide an efficient solver in the Fourier domain.
The KCF~\cite{henriques2015high} tracker further extends the method~\cite{henriques2012exploiting},
and shows improved performance can be achieved when muti-channel feature
maps are used.
Motivated by the effectiveness of the multi-channel correlation filter methods and the convolution neural network,
several methods are proposed to combine them both.
Deeply investing the representation property of different convolution layers in the CNN model,
Ma~\etal~\cite{ma2015hierarchical} propose to combine feature maps generated by three layers of convolution
filters, and introduce a coarse-to-fine searching strategy for target localization. Danelljan~\etal~\cite{danelljan2016beyond}
propose to use the continuous convolution filter for combinations of feature maps with different spatial resolutions.
As fewer model parameters are used in the model, the tracker~\cite{danelljan2016beyond} is insusceptible to
the over-fitting problem, and thus has superior performance than~\cite{ma2015hierarchical}.
Another research hotspot for the CF-based methods is how to suppress the boundary effects.
Typical methods include the trackers~\cite{danelljan2015learning} and~\cite{kiani2017learning}.
In the SRDCF tracker~\cite{danelljan2015learning},  a spatial regular term is exploited to penalize the filter
coefficients near the boundary regions. Different from~\cite{danelljan2015learning}, the BACF tracker~\cite{kiani2017learning}
directly multiplies the filter with a binary matrix. This tracker can generate real positive and negative samples for training
while at the same time share the computation efficiency of the original CF method.
Compared to our method, these trackers have not attempted to suppress the background regions inside the target bounding box,
and their learned filter weights tend to be dominated by the salient regions in the feature map.

Patch-based correlation filters have also been widely exploited~\cite{liu2015reliable,liu2016structural}.
Liu~\etal~\cite{liu2015reliable} propose an ensemble of part trackers based on the KCF method, and
use the peak-to-sidelobe ratio and the smooth constraint of confidence map for combinations of different
base trackers.
In the method~\cite{liu2016structural}, the authors attempt to learn the filter coefficients of
different patches simultaneously under the assumption that the motions of sub-patches are similar.
Li~\etal~\cite{li2015reliable} detect the reliable patches in the image, and propose to use
the Hough voting-like strategy to estimate the target states based on the sub-patches.
Most of the previous patch-based methods intend to address the problems of deformation and
partial occlusion explicitly. Different from them, our method is aimed to suppress the influence of
the non-uniform energy distribution of the feature maps and conduct a joint learning of both
discrimination and reliability.

\section{Proposed Method}
\subsection{Correlation Filter for Visual Tracking}
We first briefly revisit the conventional correlation filter (CF) formula.
Let ${\bf{y}}{\rm{ = }}{\left[ {{y_1},{y_2},...,{y_K}} \right]^\top}
\in {\mathbb{R}^{K \times 1}}$ denote gaussian shaped response, and ${\bf x}_d \in {\mathbb{R}^{K \times 1}}$
be the input vector (in the two-dimensional case, it should be a feature map) for the $d$-th channel,
then the correlation filter learns the optimal ${\bf w}$ by optimizing the following formula:
\begin{equation}
{{\bf{\dot w}}} = \arg \mathop {\min }\limits_{{{\bf{w}}}} \sum\limits_{k = 1}^K {\left( {{y_k} -
 \sum\limits_{d = 1}^D {{\bf{x}}_{k,d}^\top{{\bf{w}}_d}} } \right)_2^2}  + \lambda \left\| {{{\bf{w}}}}
 \right\|_2^2,
\label{equ:cf_formula}
\end{equation}
where ${\bf{x}}_{k,d}$ is the $k$-step circular shift of the input vector ${\bf x}_d$, ${y_k}$
is the $k$-th element of $\bf y$, ${\bf{w}}{\rm{ = }}{\left[ {{\bf{w}}_{\rm{1}}^\top,
{\bf{w}}_2^\top,...,{\bf{w}}_D^\top} \right]^\top}$ where
${{\bf{w}}_d} \in {{\mathbb{R}}^{K \times 1}}$ stands for the filter of the $d$-th channel.
For circular matrix in CF, $K$ stands for both the dimension of features and the number of training
samples.
An analytical solution can be found to efficiently solve the optimization
problem \eqref{equ:cf_formula} in the Fourier domain.

\subsection{Joint Discrimination and Reliability Modeling}
\label{eq:jdrm}

\begin{figure}[h]
  \centering
  \begin{tabular}{c}
  \includegraphics[width=0.95\linewidth,height=20mm]{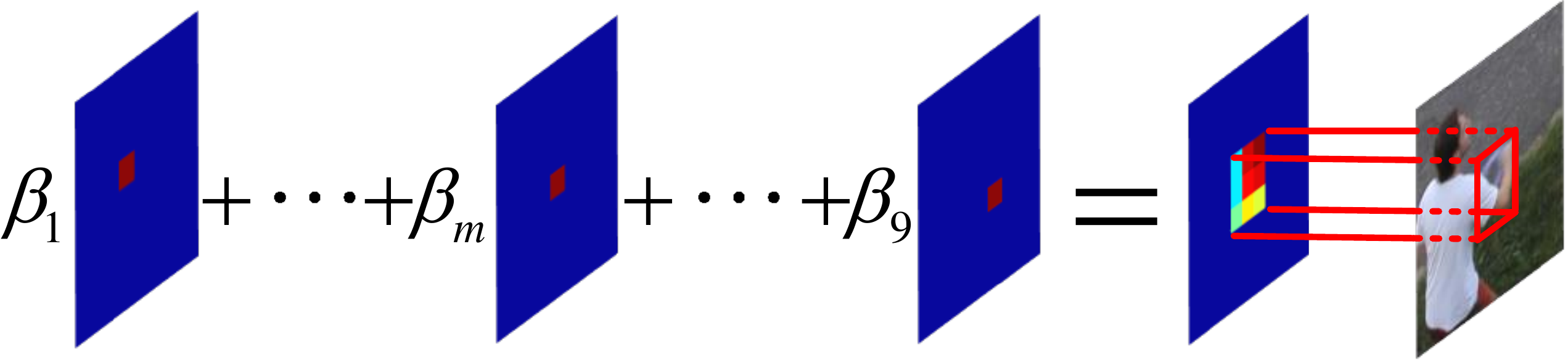}\\
  \PutCapCombine
  \end{tabular}
  \caption{Illustration showing how we compute the reliability map ${\bf v}_d$. The computed reliability map
  only has non-zeros values corresponding to the target region, thus the real positive
  and negative samples can be generated when we circularly shift the input image.}

\label{fig:combine}
\end{figure}

Different from the previous CF-based methods, we treat the filter weight ${\bf w}_d$ of the $d$-th
feature channel as the element-wise product of a base filter ${\bf h}_d$ and a reliability weight map ${\bf v}_d$,
\begin{equation}
{\bf{w}}_d = {{\bf{h}}_d} \odot {\bf v}_d,
\end{equation}
where $\odot$ is the hadamard product, ${\bf h}_d\in {\mathbb{R}}^{K \times 1}$ is used to denote
the base filter, ${\bf v}_d\in {\mathbb{R}}^{K \times 1}$ is the reliability weight for each target region,
the values of ${\bf v}_d$ corresponding to the non-target region are set to zeros (illustrated in Figure~\ref{fig:combine}).

To learn a compact reliability map, we divide the target region into $M$ patches, and use a variable $\beta_m, m\in\{1,...,M\}$
to denote the reliability for each patch ($\beta_m$ is shared across the channels), thus ${\bf v}_d$ can be written as
\vspace{-2mm}
\begin{equation}
{\bf v}_d  = \sum\limits_{m = 1}^M {{\beta _m}{{\bf{p}}_d^m}},
\label{eq:basic_assume}
\end{equation}
where ${\bf{p}}_d^m\in\mathbb{R}^{K{\times}1}$ is a binary mask (see Figure~\ref{fig:combine}) which crops the filter region
for the $m$-th patch.

Based on the previous descriptions, we attempt to jointly learn the base filter
 ${\bf h}=\left[ {\bf h}_1^\top,...,{\bf h}_D^\top\right]^\top\in \mathbb{R}^{KD{\times}1}$ and the reliability vector
 ${\boldsymbol \beta}=\left[ \beta_1,...,\beta_M  \right]^\top$ by using the following optimization problem:
 \begin{equation}
\begin{array}{l}
 \begin{array}{l}
\left[ {\dot {\bf{h}},\dot {\bm{\beta }}} \right] = \arg \mathop {\min }
\limits_{{\bf{h}},{\bm{\beta }}} f\left( {{\bf{h}},{\bm{\beta }}};{\bf X} \right)
\\
s.t.\;\;{\theta _{\min }} \le {\beta _m} \le {\theta _{\max }},\;\forall m\\
\end{array}
\label{equ:full_optimization_formula}
\end{array},
\end{equation}
where the objective function $f\left( {{\bf{h}},{\boldsymbol{\beta }}}; {\bf X} \right) $ is defined as
\begin{equation}
f\left( {{\bf{h}},{\boldsymbol{\beta }}}; {\bf X} \right) = {f_1}\left( {{\bf{h}},{\boldsymbol{\beta }}};
{\bf X} \right) + \eta {f_2}\left( {{\bf{h}}}; {\bf X} \right) + \gamma \left\| {\bf{h}} \right\|_2^2.
\label{equ:full_obj}
\end{equation}

In this equation, the first term is the data term with respect to the classification error of training samples,
the second term is a regularization term to introduce the local response consistency constraint on the filter
coefficient vector $\mathbf{h}$, and the last one is a squared $\ell_2$-norm regularization to avoid model
degradation.
In the optimization problem (\ref{equ:full_optimization_formula}), we also add some constraints on
the learned reliability coefficients $\beta_1,...,\beta_M$.
These constraints prevent all reliability weights being assigned to a small region of the target especially
when the number of training samples is limited, and encourage our model to obtain an accurate weight map.
We note that the optimization problem (\ref{equ:full_obj}) encourages learning more reliable correlation
filters (see Figure~\ref{equ:full_optimization_formula} for example).

\noindent \textbf{Data Term.}
The data term $f_1({\bf{h}},{\bm \beta};{\bf X})$ is indeed a loss function which ensures that the learned filter
has a Gaussian shaped response with respect to the circulant sample matrix.
By introducing our basic assumption in equation~\eqref{eq:basic_assume} into the
standard CF model, ${f_1}\left( {{\bf{h}},{\boldsymbol{\beta }}};{\bf X} \right) $ can
be rewritten as
\begin{equation}
\begin{array}{l}
{f_1}\left( {{\bf{h}},{\boldsymbol{\beta }}}; {\bf X} \right)\\
 = \sum\limits_{k = 1}^K {{{\left( {{y_k} - \sum\limits_{d = 1}^D {{\bf{x}}_{k,d}^\top\left( {{{\bf{v}}_d} \odot {{\bf{h}}_d}} \right)} } \right)}^2}} \\
 = \sum\limits_{k = 1}^K {{{\left( {{y_k} - \sum\limits_{d = 1}^D {{\bf{x}}_{k,d}^\top{{\bf{V}}_d}{{\bf{h}}_d}} } \right)}^2}} \\
 = \left\| {{\bf{y}} - \sum\limits_{d = 1}^D {{\bf{X}}_d^\top{{\bf{V}}_d}{{\bf{h}}_d}} } \right\|_2^2\\
 = \left\| {{\bf{y}} - {{\bf{X}}^\top}{\bf{Vh}}} \right\|_2^2
\end{array},
\label{equ:f1}
\end{equation}
where ${{\bf{V}}_d} = {\rm diag}\left( {{\bf{v}}_d}\left( 1 \right),{{\bf{v}}_d}\left( 2 \right),...,{{\bf{v}}_d}
\left( K \right) \right) \in {\mathbb{R}^{K \times K}}$ is a diagonal matrix,
${{\bf{X}}_d} = \left[ {{{\bf{x}}_{1,d}},{{\bf{x}}_{2,d}},...,{{\bf{x}}_{K,d}}} \right] \in {\mathbb{R}^{K \times K}}$
is the circulant matrix of the $d$-th channel, ${\bf{X}} = {\left[ {{\bf{X}}_1^\top,{\bf{X}}_2^\top,...,{\bf{X}}_D^\top}
\right]^\top}\in \mathbb{R}^{KD{\times}K}$ stands for a contactated matrix of all circulant matrices from different channels,
and ${\bf{V}} = {\bf{V}}_1 \oplus {\bf{V}}_2 \oplus  \cdot  \cdot  \cdot  \oplus {\bf{V}}_D \in {\mathbb{R}^{DK \times DK}}$
denotes a block diagonal matrix where ${\bf{V}}_d$ is the $d$-the diagonal block.

\vspace{2mm}
\noindent \textbf{Local Response Consistency.} The regularization term ${f_2}\left( {{\bf{h}}};{\bf X} \right)$ constrains that the base filter generates consistent responses for
different fragments of the cyclically shifted sample. By this means, the base filter learns equal importance for
each local region, and reliability information is separated from the base filter. The term ${f_2}\left( {{\bf{h}}};{\bf X} \right)$ can be defined as
\begin{equation}
\begin{array}{l}
{f_2}\left( {{\bf{h}}};{\bf X} \right)\\
 = {\sum\limits_{k = 1}^K {\sum\limits_{m,n}^M {\left( {\sum\limits_{d = 1}^D {{{\left( {{\bf{P}}_d^m{{\bf{x}}_{k,d}}} \right)}^\top}{{\bf{h}}_d}}  - \sum\limits_{d = 1}^D {{{\left( {{\bf{P}}_d^n{{\bf{x}}_{k,d}}} \right)}^\top}{{\bf{h}}_d}} } \right)} } ^2}\\
 = \sum\limits_{m,n}^M {\left\| {\sum\limits_{d = 1}^D {{\bf{X}}_d^\top{\bf{P}}_d^m{{\bf{h}}_d}}  - \sum\limits_{d = 1}^D {{\bf{X}}_d^\top{\bf{P}}_d^n{{\bf{h}}_d}} } \right\|_2^2} \\
 = \sum\limits_{m,n}^M {\left\| {{{\bf{X}}^\top}{{\bf{P}}^m}{\bf{h}} - {{\bf{X}}^\top}{{\bf{P}}^n}{\bf{h}}} \right\|_2^2}
\end{array},
\label{equ:f2}
\end{equation}
where ${\bf{P}}_d^m = {\rm{diag(}}{\bf{p}}_d^m{\rm{(1),}}{\bf{p}}_d^m{\rm{(2),}}...{\rm{,}}{\bf{p}}_d^m{\rm{(}}K{\rm{))}}\in \mathbb{R}^{K{\times}K}$, ${{\bf{P}}^m} = {\bf{P}}_1^m \oplus {\bf{P}}_2^m \oplus  \cdot  \cdot  \cdot  \oplus {\bf{P}}_D^m \in {\mathbb{R}^{DK \times DK}}$. For each cyclically shifted sample ${\bf x}_{k,d}$, ${{{\left( {{\bf{P}}_d^m{{\bf{x}}_{k,d}}} \right)}^\top}{{\bf{h}}_d}}$ is the response for the $m$-th fragment of ${\bf x}_{k,d}$.

\subsection{Joint Discrimination and Reliability Learning}
\label{eq:jdrl}
Based on the discussions above, the base filter and the reliability vector can be jointly learned by solving
the optimization problem (\ref{equ:full_optimization_formula}), which is a non-convex but differentiable
problem for both ${{\bf h}}$ and ${\boldsymbol \beta}$. However, it can be converted into a convex
differentiable problem if either  ${{\bf h}}$ or ${\boldsymbol \beta}$ is known.
Thus, in this work, we attempt to solve the optimal ${\dot {\bf h}}$ and $\dot {\bm \beta}$ via the alternating
direction method.

\vspace{2mm}
\noindent  {\bf Solving ${{\bf h}}$}.
To solve the optimal ${{\bf h}}$, we first compute the derivative of the objective function (\ref{equ:full_obj}) ,
then by setting the derivative to be zero, we obtain the following normal equations:
\begin{equation}
{\bf{Ah}} = {\bf{VXy}}.
 \label{equ:normal_equations}
\end{equation}

The matrix $\bf A$ is defined as
\begin{equation}
  \begin{array}{l}
{\bf{A}} = {{\bf g}({\bf{V}},{{\bf{X}}})}  + 2\eta{\sum\limits_{m = 1}^M {M{\bf g}({{\bf{P}}^m},{{\bf{X}}})} }
\\
{\kern 3mm}- 2{\eta}{{\bf g}(\sum\limits_{m = 1}^M {{{\bf{P}}^m},{{\bf{X}}}})}+{\gamma}{\bf I}
\end{array},
\label{equ:definition_A}
\end{equation}
where ${\bf{g}}\left( {{\boldsymbol{\Lambda }},{\bf{R}}} \right) =
{{\boldsymbol{\Lambda }}^\top}{\bf{R}}{{\bf{R}}^\top}{\bf{\Lambda }}$,
$\bf{R}$ is a circulant matrix and ${\boldsymbol{\Lambda }}$ is a diagonal matrix.

In this work, we exploit the conjugate gradient descent method due to its fast convergence speed.
The update process can be performed via the following iterative steps~\cite{NoceWrig06}:
\begin{equation}
\begin{array}{l}
{\alpha ^{(i)}} = {{\bf{r}}^{(i)}}_{}^{\top}{{\bf{r}}^{(i)}}/{{\bf{u}}^{(i)}}^{\top}{\bf{A}}{{\bf{u}}^{(i)}}\\
{{\bf{h}}^{(i + 1)}} = {{\bf{h}}^{(i)}} + {\alpha ^{(i)}}{{\bf{u}}^{(i)}}\\
{{\bf{r}}^{(i + 1)}} = {{\bf{r}}^{(i)}} + {\alpha ^{(i)}}{\bf{A}}{{\bf{u}}^{(i)}}\\
{\mu ^{(i+1)}} = \left\| {{{\bf{r}}^{(i + 1)}}} \right\|_2^2/\left\| {{{\bf{r}}^{(i)}}} \right\|_2^2\\
{{\bf{u}}^{(i + 1)}} =  -{{\bf{r}}^{(i + 1)}} + {\mu ^{(i+1)}}{{\bf{u}}^{(i)}}
\end{array},
\end{equation}
where ${\bf u}^{(i)}$ denotes the search direction at the $i$-th iteration, ${\bf r}^{(i)}$ is the residual after
the $i$-th iteration.
Clearly, the computational load lies in the update of ${\alpha ^{(i)}}$ and ${\bf r}^{(i+1)}$ since it requires
to compute ${{\bf{u}}^{(i)}}^\top{\bf{A}}{{\bf{u}}^{(i)}}$ and ${\bf{A}}{{\bf{u}}^{(i)}}$ in each iteration.
As shown in equation~\eqref{equ:definition_A}, the first three terms have the same form.
For clarity, we take the first term as an example to show how we compute ${{\bf{u}}^{(i)}}^\top{\bf{A}}{{\bf{u}}^{(i)}}$
and ${\bf{A}}{{\bf{u}}^{(i)}}$ efficiently.
Let ${\bf A}_1$ denote the first term of equation~\eqref{equ:definition_A}, then
\begin{equation}
\begin{array}{l}
{{\bf{u}}^{(i)}}^\top{{\bf{A}}_1}{{\bf{u}}^{(i)}} = {{\bf{u}}^{(i)}}^\top{{\bf{V}}^\top}{\bf{X}}{{\bf{X}}^\top}{\bf{V}}{{\bf{u}}^{(i)}}\\
{\kern 52pt}  = \sum\limits_{d = 1}^D {\left\| {{\bf{X}}_d^\top{{\bf{V}}_d}{\bf{u}}_d^{(i)}} \right\|_2^2} \\
{\kern 52pt}   = \frac{1}{{{K}}}\sum\limits_{d = 1}^D {\left\| {\widehat {\bf{X}}_d^H \odot \mathcal{F}\left( {{{\bf{V}}_d}{\bf{u}}_d^{(i)}} \right)} \right\|_2^2}
\end{array},
\label{equ:kak}
\end{equation}
where ${\bf{\widehat X}}_{d}=\mathcal{F}({\bf x}_{d})$ is the Fourier transform of the base vector ${\bf x}_{d}$
(corresponding to the input image without shift),  ${{\bf{u}}^{(i)}_d}$ is the subset of ${\bf{u}}^{(i)}$ corresponding
to the $d$-th channel, $(\cdot)^H$ denotes the conjugate of a vector.
Because ${\bf{V}}_d{{\bf{u}}^{(i)}_d}$ is a vector and ${\bf{X}}_{d}$ is the circulant matrix,
the operation ${\bf{X}}_{d}^{\top}({\bf{V}}_d{{\bf{u}}_d^{(i)}})$ can be viewed as a circular
correlation process and can be efficiently computed in the Fourier domain.

Similarly, ${{\bf A}_1}{\bf u}^{(i)}$ can be computed as
\begin{equation}
{{\bf{A}}_1}{{\bf{u}}^{(i)}} = {{\bf{V}}^\top}{\bf{X}}{{\bf{X}}^\top}{\bf{V}}{\bf u}^{(i)} = \left[ \begin{array}{l}
{\bf{V}}_1^\top{{\bf{X}}_1}\sum\limits_{j = 1}^D {{\bf{X}}_j^\top{{\bf{V}}_j}{\bf{u}}_j^{(i)}} \\
{\bf{V}}_2^\top{{\bf{X}}_2}\sum\limits_{j = 1}^D {{\bf{X}}_j^\top{{\bf{V}}_j}{\bf{u}}_j^{(i)}} \\
...\\
{\bf{V}}_D^\top{{\bf{X}}_D}\sum\limits_{j = 1}^D {{\bf{X}}_j^\top{{\bf{V}}_j}{\bf{u}}_j^{(i)}}
\end{array} \right].
\label{equ:ak}
\end{equation}

The $d$-th term ${\bf{V}}_d^\top{{\bf{X}}_d}\sum\limits_{j = 1}^D {{\bf{X}}_j^\top{{\bf{V}}_j}{\bf{u}}_j^{(i)}}$
can be computed as
\begin{equation}
\begin{array}{l}
{\bf{V}}_d^\top{{\bf{X}}_d}\sum\limits_{j = 1}^D {{\bf{X}}_j^\top{{\bf{V}}_j}{\bf{u}}_j^{(i)}} \\
 = {\bf{V}}_d^\top{\mathcal{F}^{ - 1}}\left( {{{{\bf{\widehat X}}}_d} \odot \sum\limits_{j = 1}^D
 {{\bf{\widehat X}}_j^H \odot \mathcal{F}({{\bf{V}}_j}{\bf{u}}_j^{(i)})} } \right)
\end{array},
\end{equation}
where ${{\bf{\widehat X}}_{j}^H \odot \mathcal{F}({{\bf{V}}_j}{\bf{u}}_j^{(i)})}$ has been computed in
equation~\eqref{equ:kak} and can be directly used. The computational complexities of \eqref{equ:kak} and
\eqref{equ:ak} are therefore $\mathcal{O}(DK\log K)$.

\vspace{2mm}
\noindent  {\bf Solving ${{\bm \beta}}$}.
If the filter vector $\mathbf{h}$ is given, the reliability weight vector
$\boldsymbol{\beta}  = {\left[ {{\beta _1},{\beta _2},...,{\beta _M}} \right]^\top}$ can be obtained
by solving the following optimization problem:
\begin{equation}
\begin{array}{l}
{\boldsymbol{\dot \beta }} = \arg \mathop {\min }\limits_{\bm{\beta }} \left\| {{\bf{y}} - \sum\limits_{d = 1}^D
{{\bf{X}}_d^\top{{\bf{V}}_d}{{\bf{h}}_d}} } \right\|_2^2\\
s.t.\;\;\;{\theta _{\min }} \le {\beta _m} \le {\theta _{\max }},\;\forall m
\end{array},
\label{equ:optimize_B}
\end{equation}
where the term ${f}_2(\bf h;{\bf X} )$ is ignored as it does not include $\bm \beta$.
With some derivations,
the problem (\ref{equ:optimize_B}) can be converted as follows:
\begin{equation}
\begin{array}{l}
{\bm{\dot \beta}} = \arg \mathop {\min }\limits_{{\bm \beta}}{{\bm{\beta }}^\top} {{\bf{C}}^\top{{\bf{C}}}} {\bm{\beta }} - 2{{\bm{\beta }}^\top} {\bf{C}}^\top {\bf y}\\
s.t.{\kern 19pt} {\theta _{\min }} < {\beta _m} < {\kern 1pt} {\theta _{\max }}, {\kern 1mm} \forall m
\end{array},
\label{equ:solve_beta}
\end{equation}
where ${\bf{C}}=\left[ {\bf{C}}^1,...,{\bf{C}}^M   \right]\in \mathbb{R}^{K{\times}M}$,
and ${\bf{C}}^m$ is computed as ${\bf{C}}^m =  {{\mathcal{F}^{ - 1}}} (\sum\limits_{d = 1}^D
{\bf{\widehat X}}_{d}^H \odot \mathcal{F}({\bf{P}}_d^m{{\bf{h}}_d}))$, whose computational
complexity is $\mathcal{O}(DK{\rm log}(K))$.
This optimization problem is a convex quadratic programming method, which can be efficiently
solved via standard quadratic programming.

\subsection{Model Extension}
We note the proposed model (Section~\ref{eq:jdrm}) and its optimization method (Section~\ref{eq:jdrl})
are defined and derived based on the training sample in one frame.
Recent studies (like~\cite{danelljan2016beyond}) demonstrate that it is more effective to
learn the correlation filter using a set of training samples from multiple frames.
Thus, we can extend our optimization problem (\ref{equ:full_optimization_formula})
to consider multi-frame information as follows:
\begin{equation}
\begin{array}{l}
\left[ {{\bf{\dot h}},{\bm{\dot \beta }}} \right] = \arg \mathop {\min }\limits_{{\bf{h}},{\bm{\beta }}} \sum\limits_{t = 1}^T
{{\kappa _t}f({\bf{h}},{\bm{\beta }};{{\bf{X}}^t})} \\
s.t. {\kern 8pt} {\theta _{\min }} < {\beta _m} < {\theta _{\max }},{\kern 4pt} {\kern 1pt} \forall m
\end{array},
\label{eq:newopt}
\end{equation}
where ${\bf X}^t$ denotes the sample matrix in the $t$-th frame, ${\kappa _t}$ is a temporal weight
to indicate the contribution of the $t$-th frame.
It is not difficult to prove that the previous derivations (in Section~\ref{eq:jdrm} and~\ref{eq:jdrl})
can be applicable for solving  the optimization problem (\ref{eq:newopt}).

\begin{figure}
  \centering
  \includegraphics[width=1\linewidth]{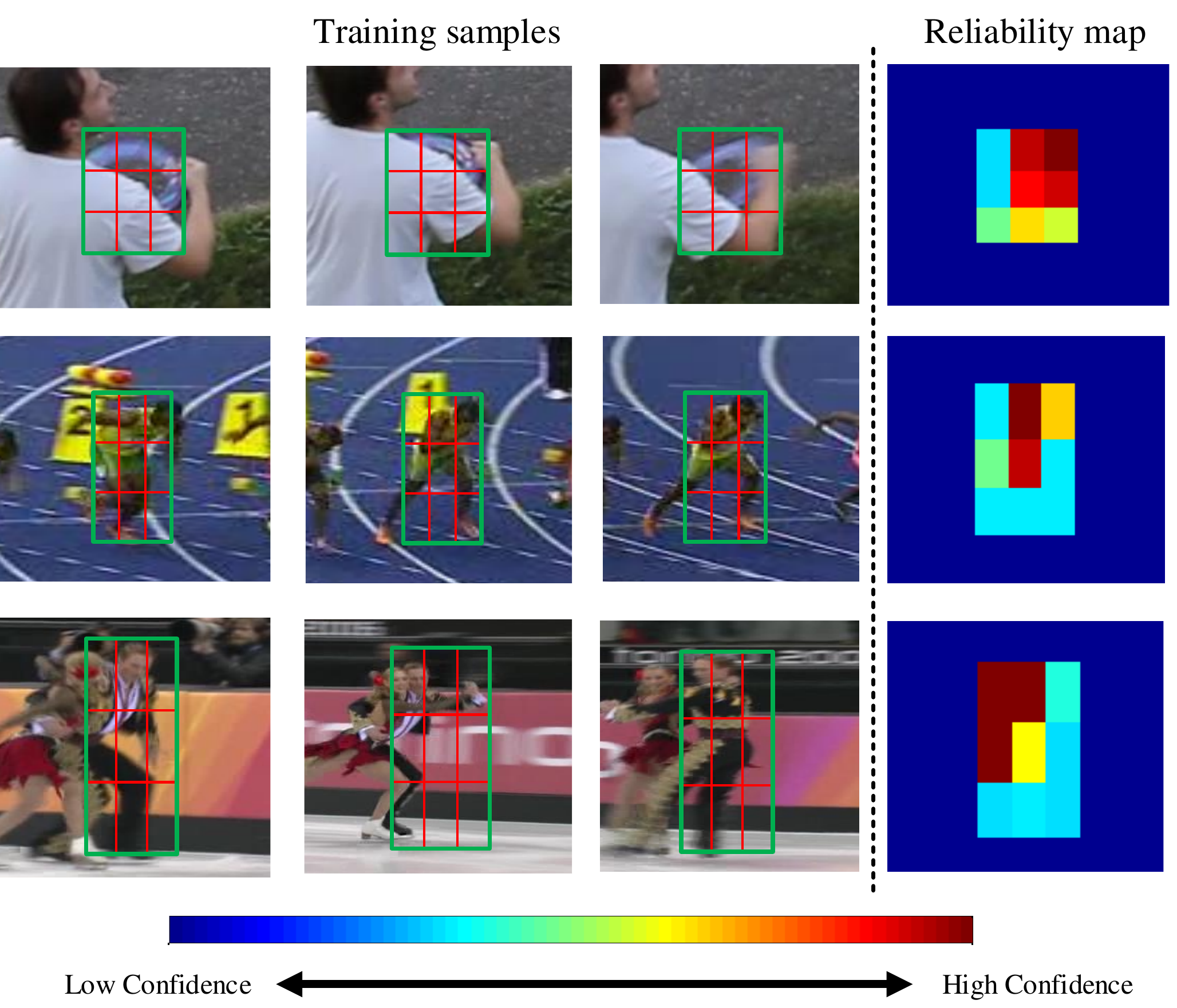}\\
  \caption{Illustration showing that reliable regions can be determined through the joint learning formula. We show three example training samples on the left three columns, and show
  the learned reliable weight maps on the fourth column. }
  \label{fig:show_reliabiligy}
\end{figure}

In Figure~\ref{fig:show_reliabiligy}, we provide examples showing that our tracker can accurately
learn the reliability value for each patch region.
In the first row, the left part of the frisbee is frequently occluded, our method learns a small reliability
value for such regions.
The example in the second row demonstrates that our method can accurately determine that the fast moving
legs are not reliable.
In the last example, the background regions are associated with small weights via the proposed model,
thereby facilitating the further tracking process.

\section{Model Update}
Most correlation filter based tracking algorithms perform model update in each frame, which results in high computation load. Recently,
the ECO method proves that the sparse update mechanism is a better choice for the CF based trackers.
Following the ECO method, we also exploit the sparse update mechanism in our tracker. In the update frame, we use the conjugate gradient descent method to update the
base filter coefficient vector $\bf h$ and then we update ${\bm \beta}$ based on the updated base filter by solving a quadratic programming problem.
In each frame, we initialize the weight for the training frame as $\omega$ and weights of previous training samples are decayed as $(1-\omega)\kappa _t$.
When the number of training samples exceeds the pre-defined value $T_{\rm max}$, we follow the ECO method and use the
Gaussian Mixture Model (GMM) for sample fusion.
 We refer the readers to~\cite{danelljan2016eco} for more details.

\begin{figure*}[t]
  \centering
  \begin{tabular}{c@{}c@{}c@{}c}
  \includegraphics[width=0.2445\linewidth,height=35mm]{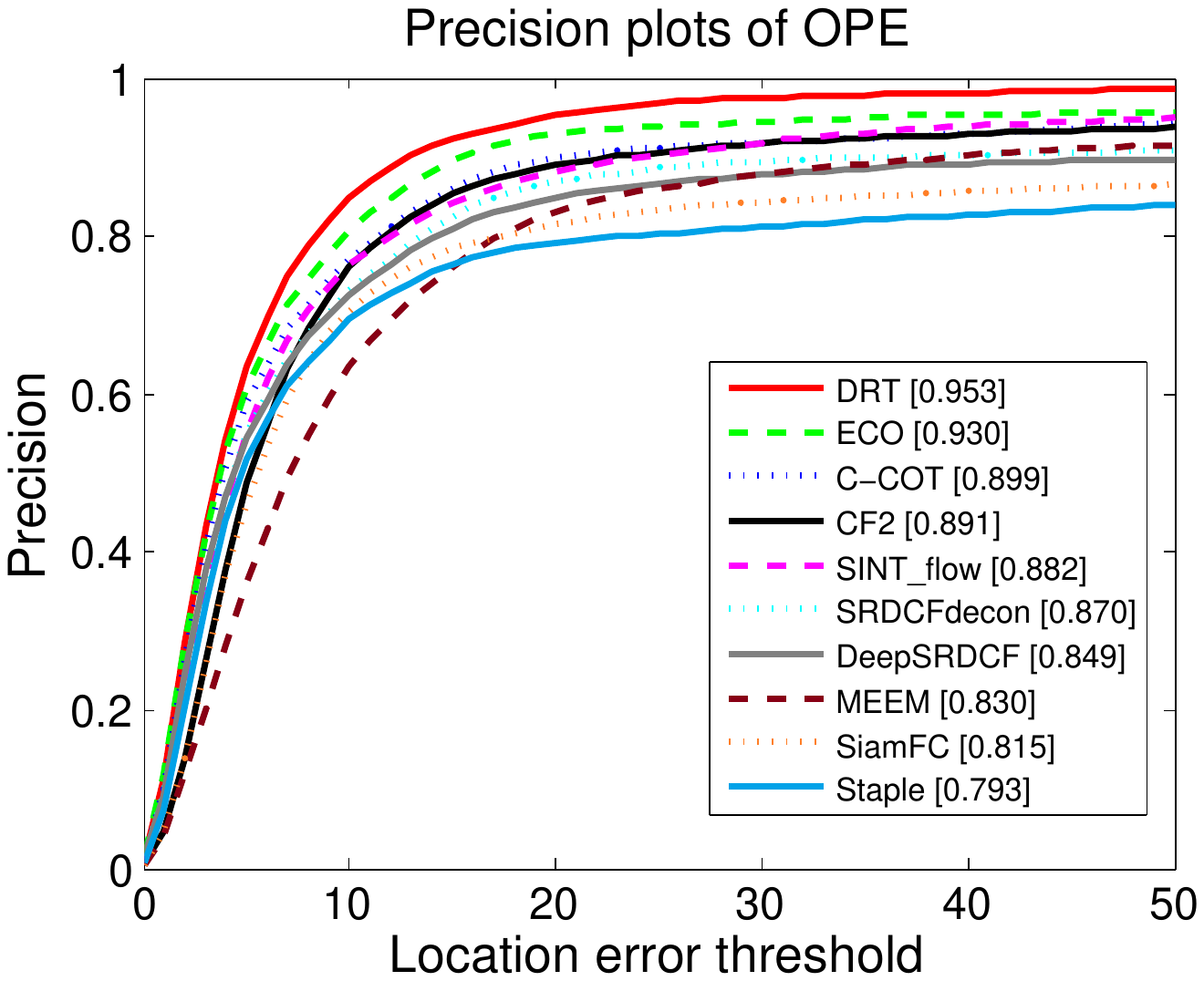}
  \ &
  \includegraphics[width=0.2445\linewidth,height=35mm]{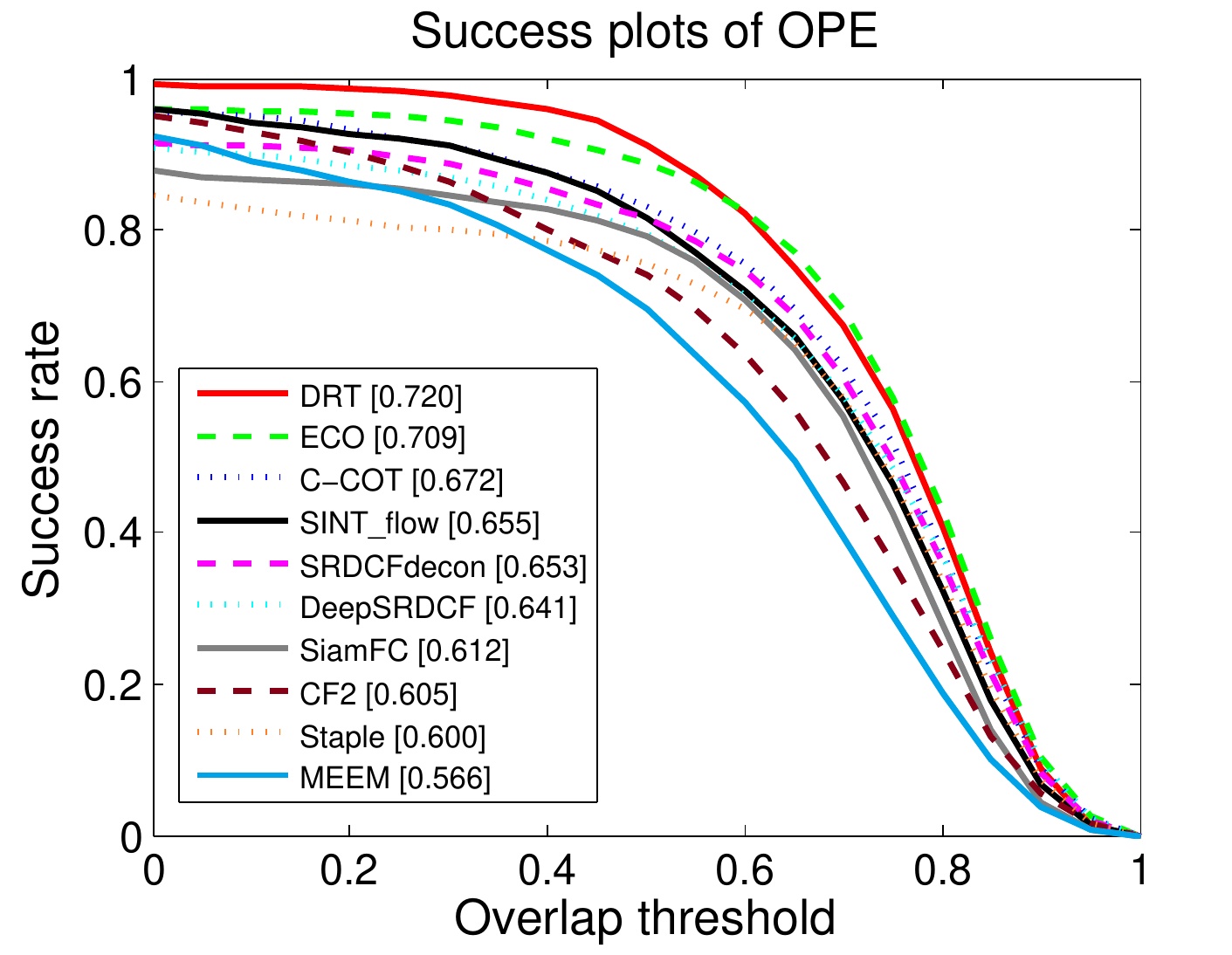}
  \ &
   \includegraphics[width=0.2445\linewidth,height=35mm]{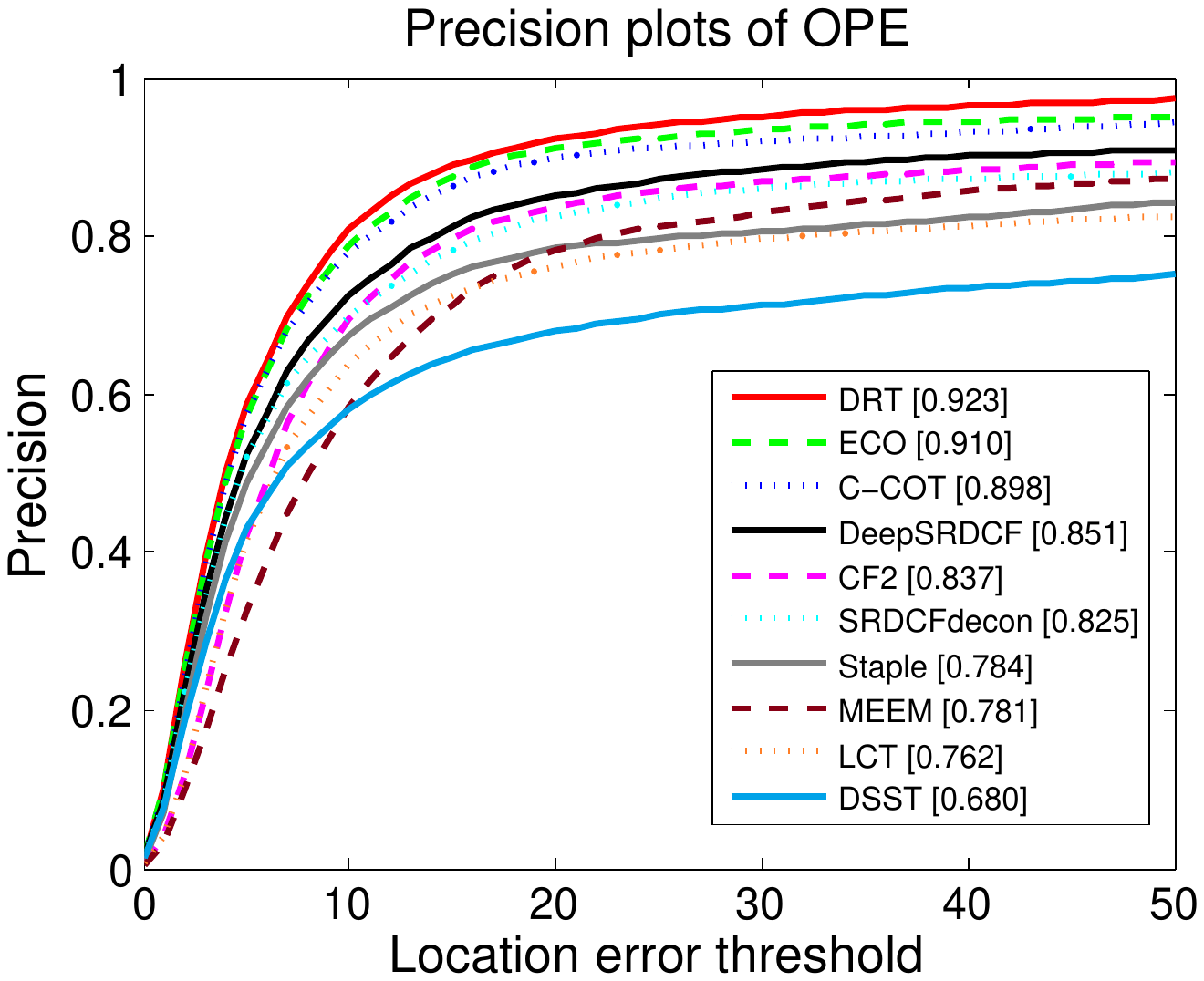}
  \ &
  \includegraphics[width=0.2445\linewidth,height=35mm]{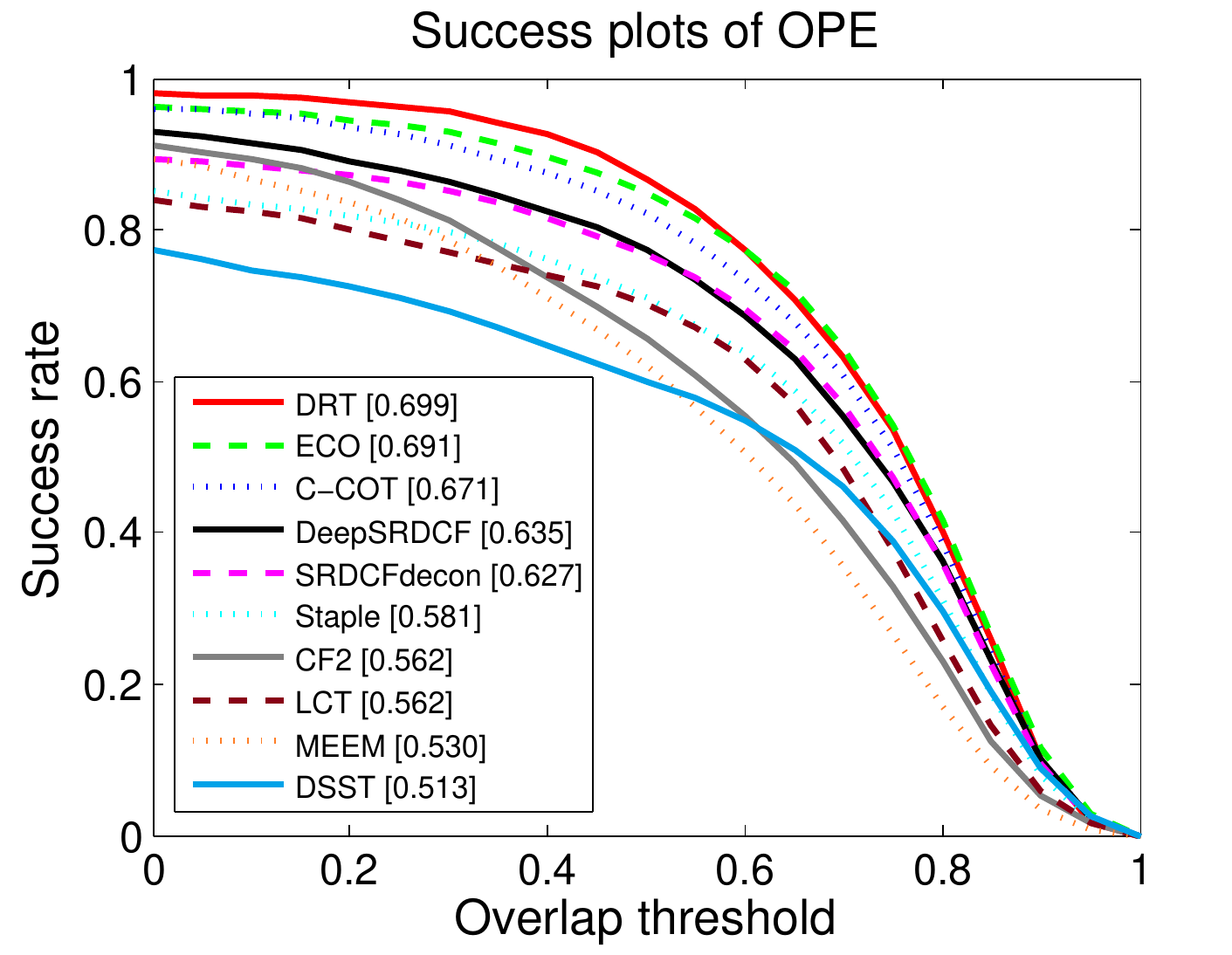}
  \\
  \PutCapDataset
  \end{tabular}

  \caption{Precision and success plots of different trackers on the OTB-2013 and OTB-2015 datasets in terms of the OPE rule.
  This figure only shows the plots of top 10 trackers for clarity.
  In the legend behind of name of each tracker, we show the distance precision score at the threshold on 20 pixels and the area under curve (AUC) score.}
 \label{fig:otb}
\end{figure*}

\begin{figure*}[http]
\centering
\begin{tabular}{c@{}c@{}c@{}c}
\includegraphics[width=0.2445\linewidth,height=34.5mm]{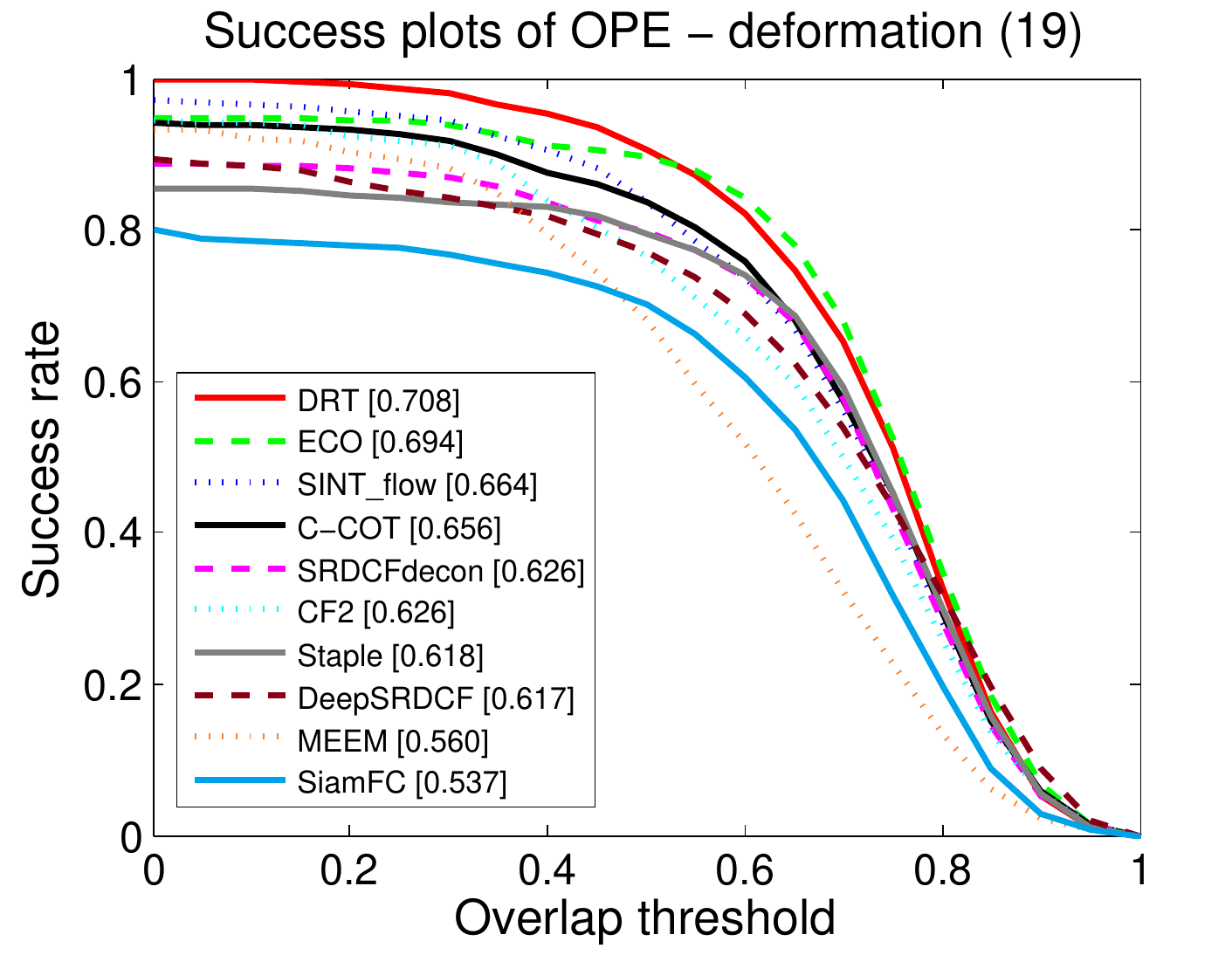}
\ &
\includegraphics[width=0.2445\linewidth,height=34.5mm]{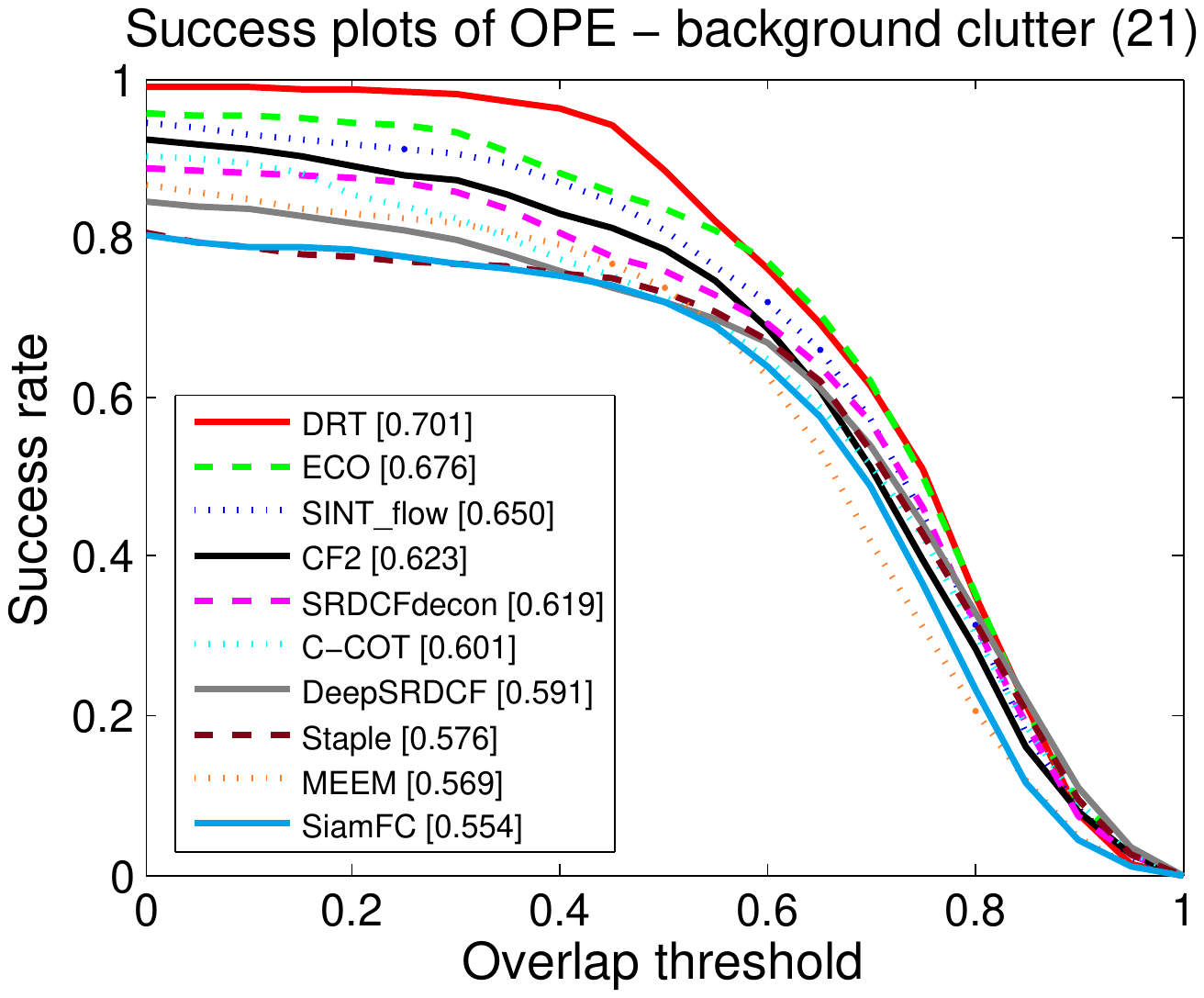}
\ &
\includegraphics[width=0.2445\linewidth,height=34.5mm]{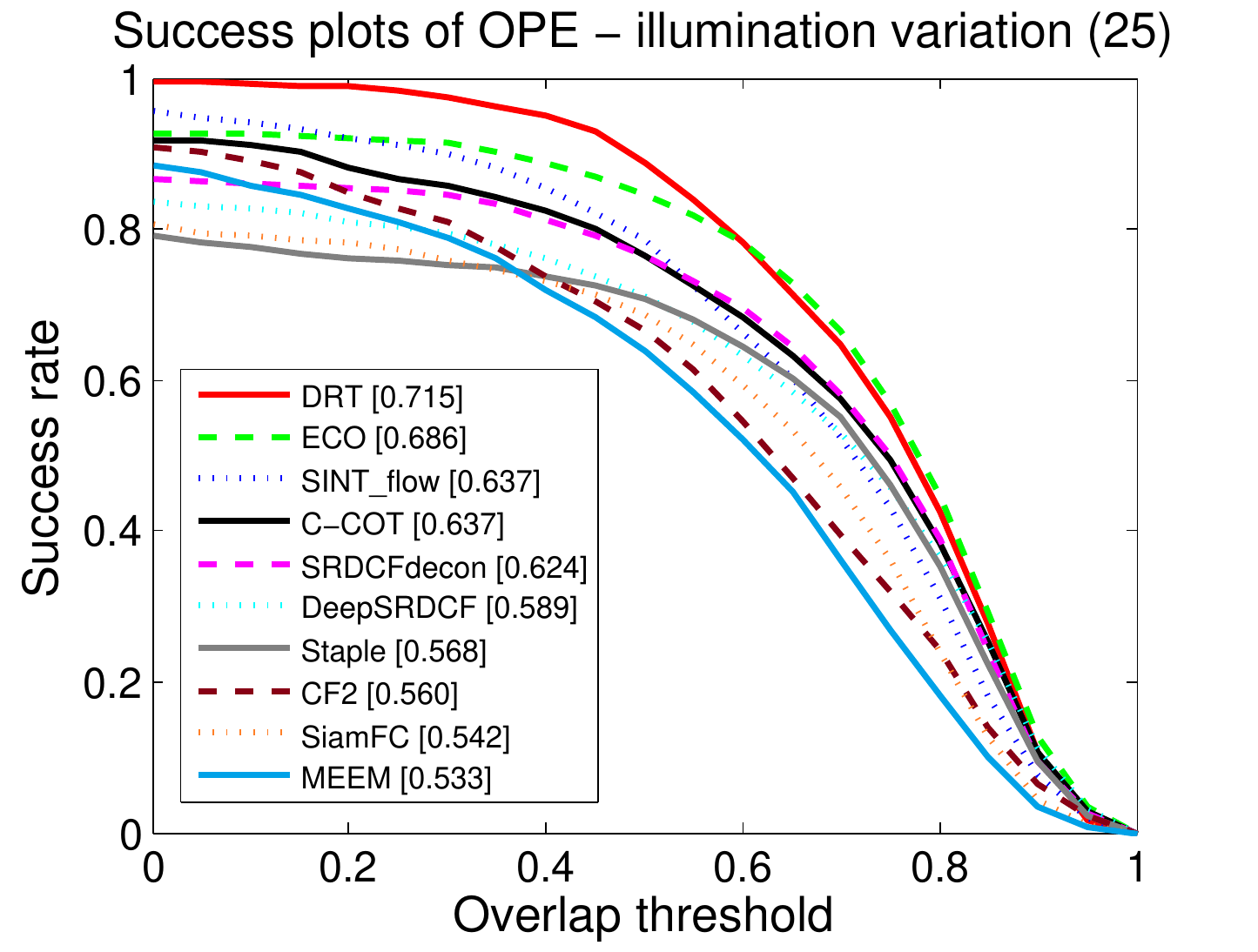}
\ &
\includegraphics[width=0.2445\linewidth,height=34.5mm]{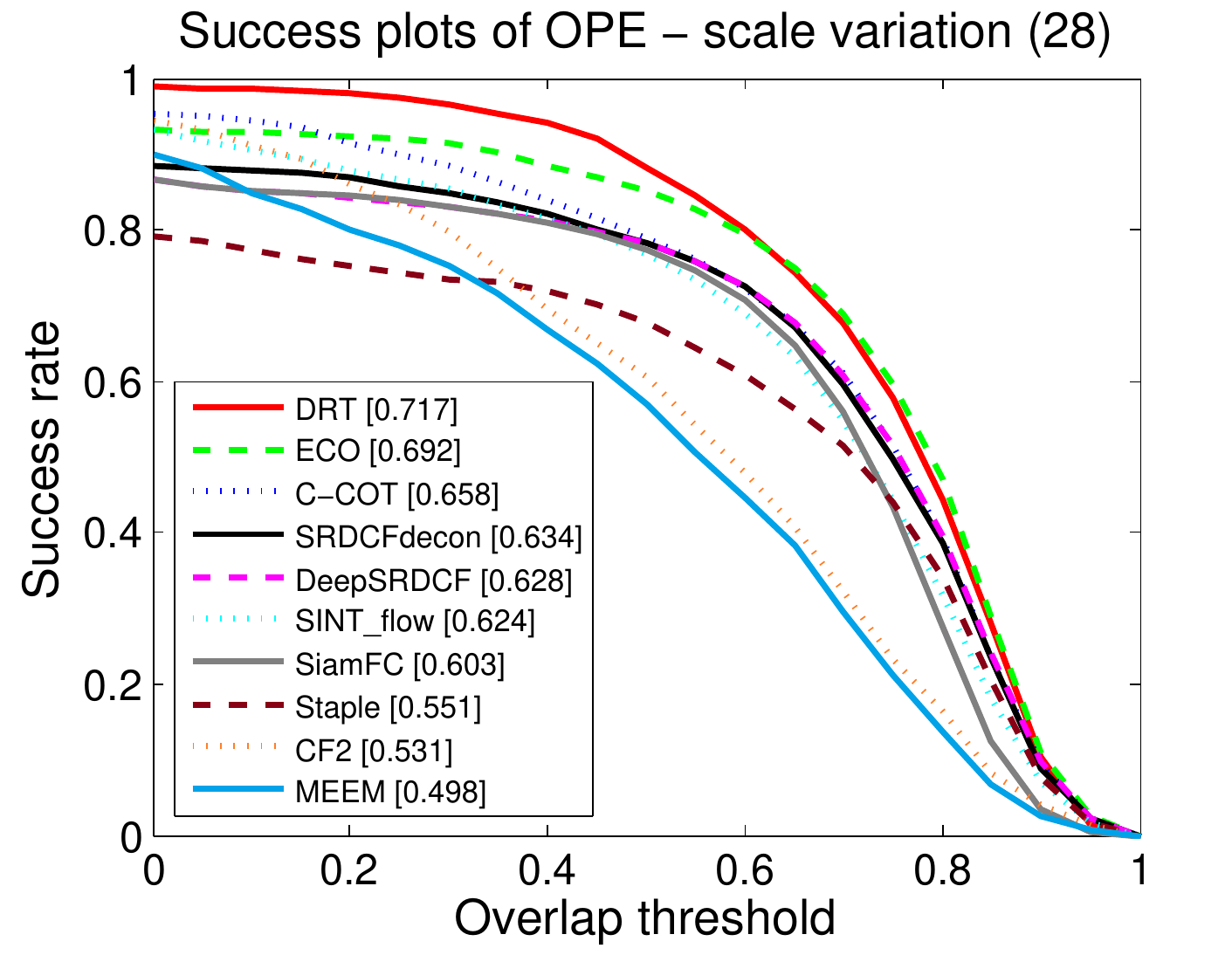}
\ \\
\includegraphics[width=0.2445\linewidth,height=34.5mm]{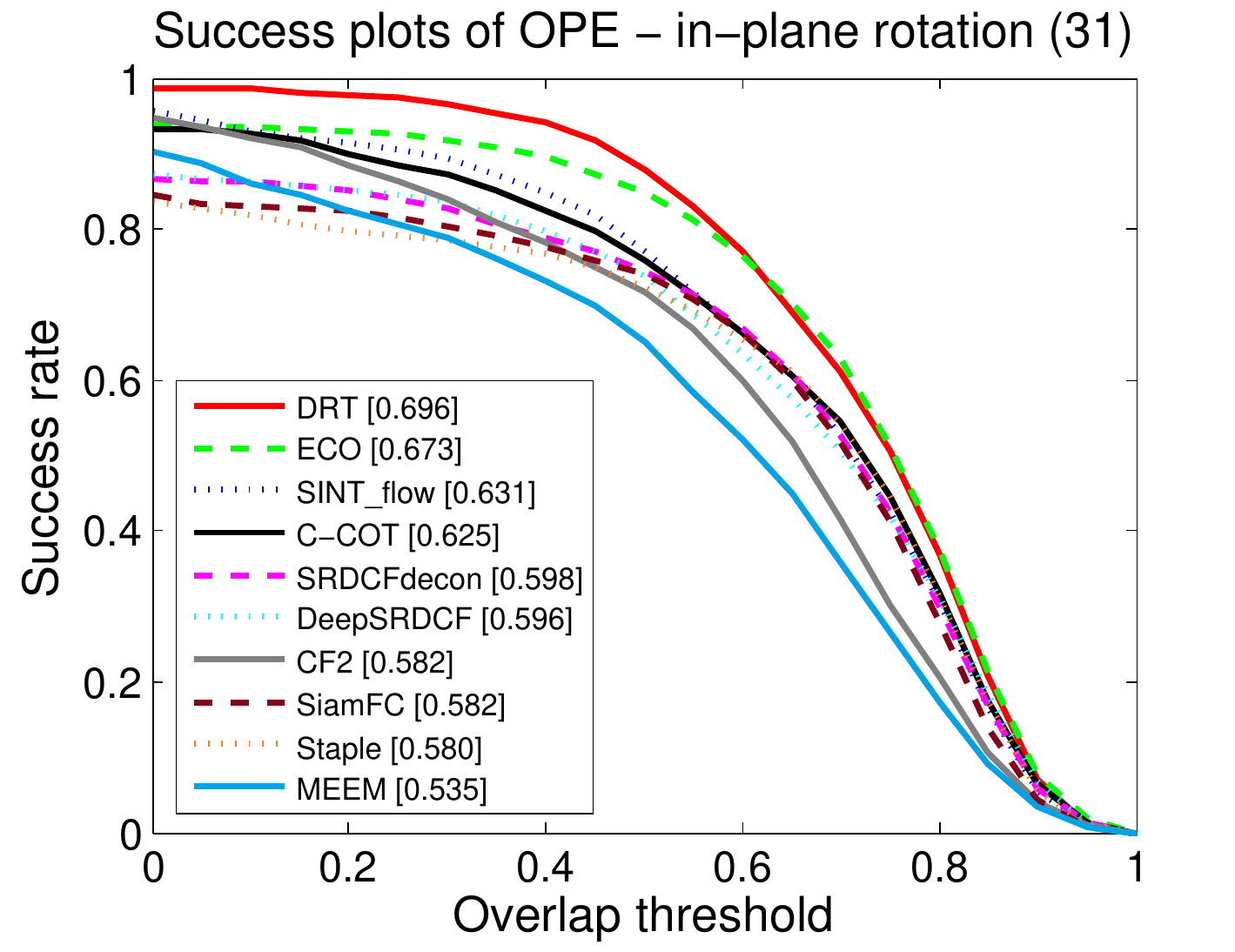}
\ &
\includegraphics[width=0.2445\linewidth,height=34.5mm]{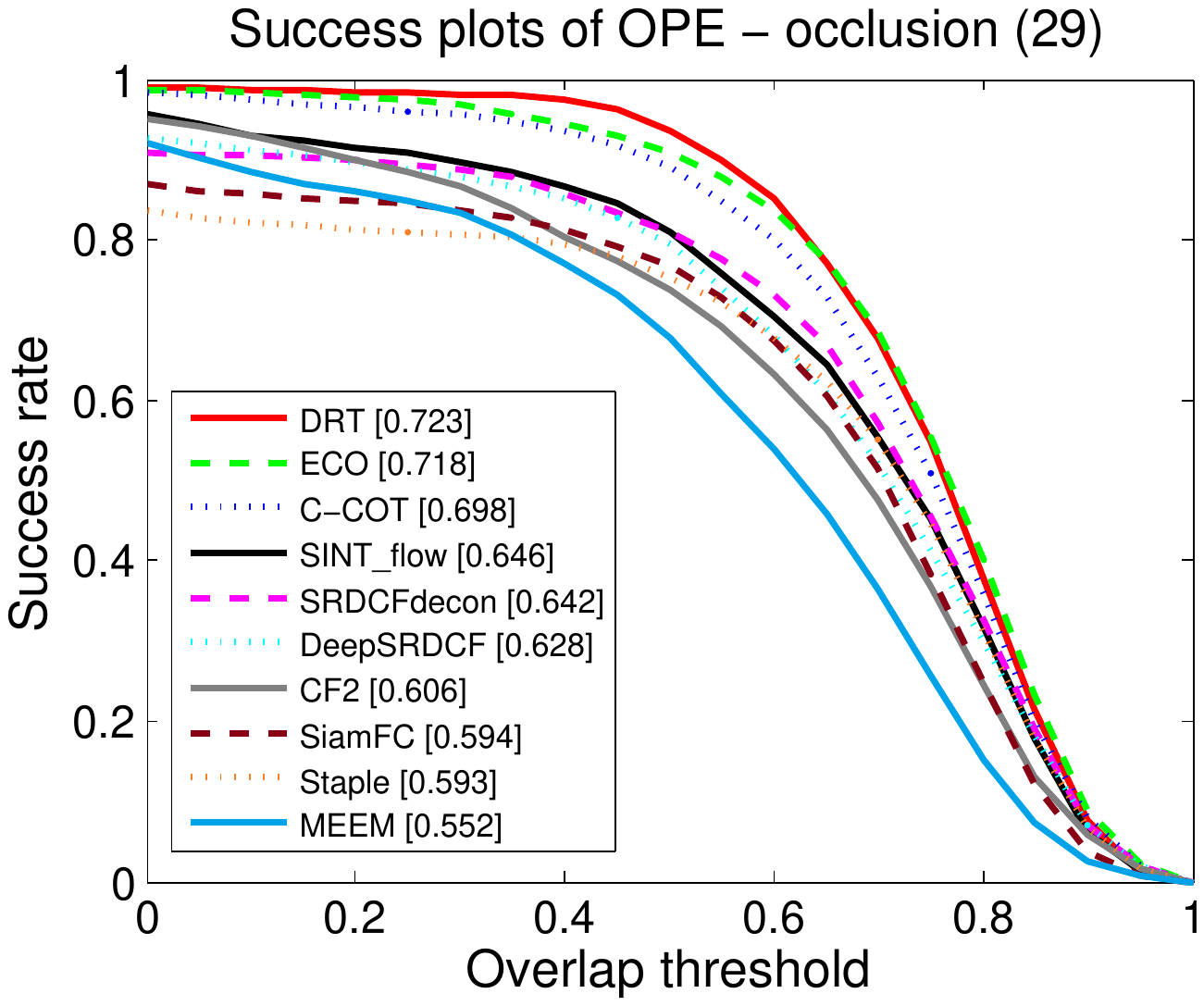}
\ &
\includegraphics[width=0.2445\linewidth,height=34.5mm]{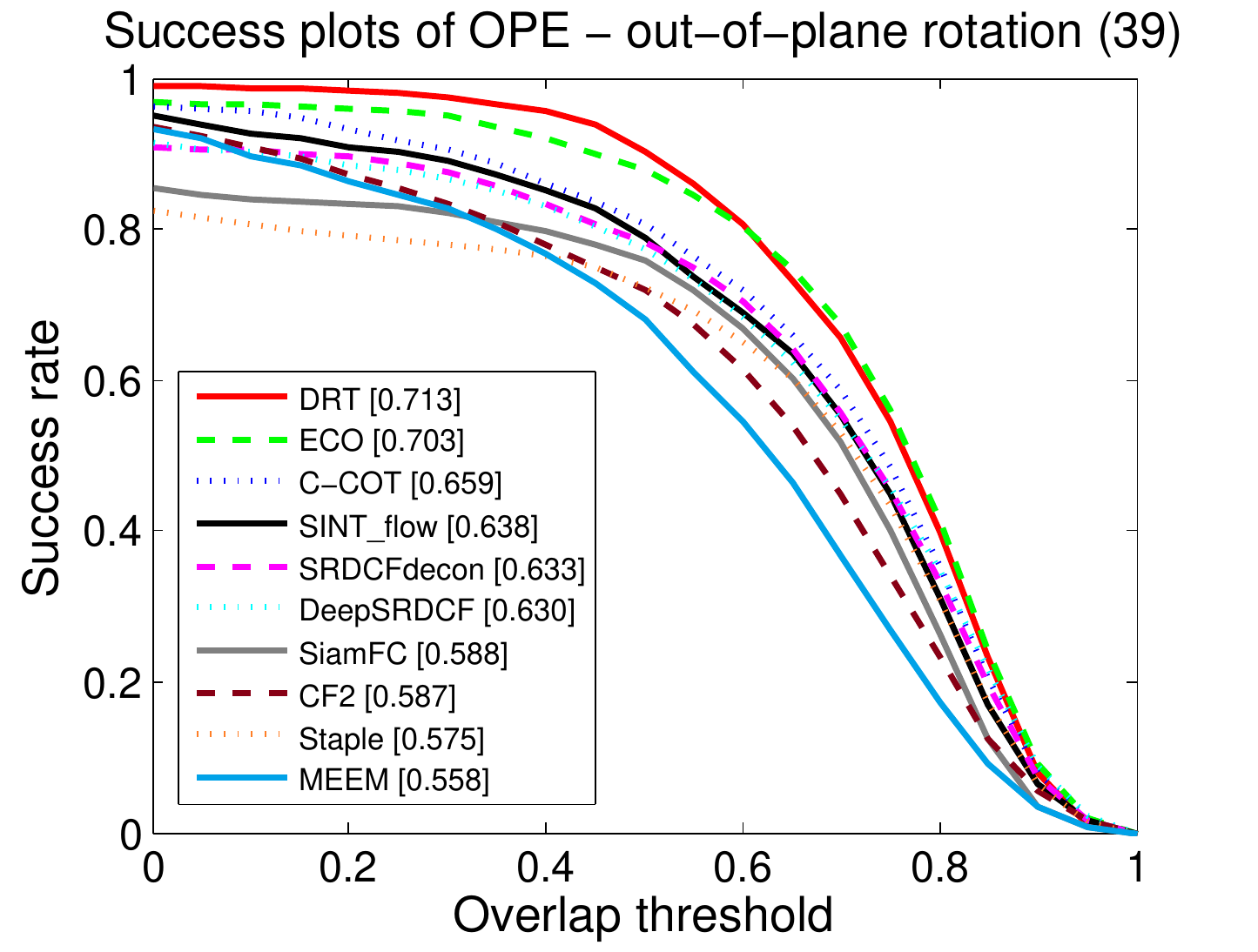}
\ &
\includegraphics[width=0.2445\linewidth,height=34.5mm]{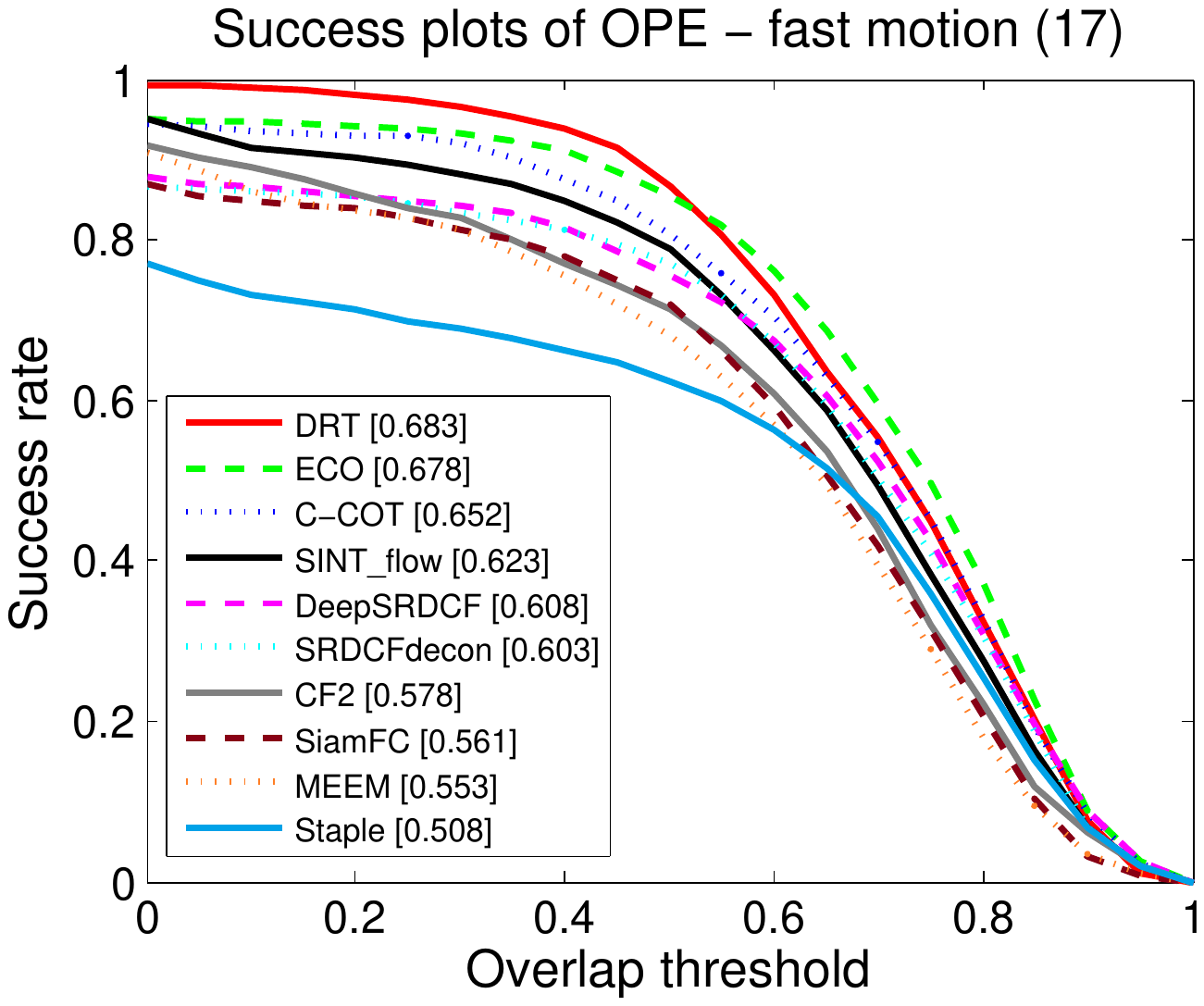}
\end{tabular}
\caption{Performance evaluation on different attributes of the benchmark
in terms of the OPE criterion. Merely top 10 trackers for each attributes are illustrated for clarity.}
\label{fig:attributes}
\vspace{-2mm}
\end{figure*}

\section{Target Localization}
In the detection process at the $t$-th frame, we use a multi-scale search strategy~\cite{danelljan2016beyond,danelljan2016eco} for joint target localization and scale estimation.
We extract the ROI regions with different scales centred in the estimated position of last frame, and obtain the multi-channel feature map ${\bf x}_d^s, d=\{1,...,D\}, s=\{1,...,S\}$ for the ROI region, where $s$ is the scale index.
Then we compute the response for the target localization in scale $s$ as
\begin{equation}
{\bf r}_s = \sum\limits_{d = 1}^D {{\mathcal{F}^{ - 1}}(\mathcal{F}({{\bf{w}}_d}) \odot (\mathcal{F}({{\bf{x}}_d^s}))^H  )}.
\end{equation}

The target location and scale are then jointly determined by finding the maximum value in the $S$ response maps. This joint estimation strategy
shows better performance than the previous methods, which first estimate the target position and then refine the scale based on the estimated
position.

\begin{table*}[http]
\caption{Performance evaluation of different state-of-the-art trackers in the
VOT-2016 dataset. In this dataset, we compare our DRT method
with the top 10 trackers. The best two results are marked in red and blue bold fonts,
respectively.}
\label{tab:table1}
\vspace{0mm}
\small
\begin{center}
\label{tab1}
\begin{tabular}{cccccccccccc}
\hline
\hline
&\textbf{STAPLE+}&\textbf{SRBT}&\textbf{EBT}&\textbf{DDC}&\textbf{Staple}&\textbf{MLDF}
&\textbf{SSAT}&\textbf{TCNN}&\textbf{C-COT}&\textbf{ECO}&\textbf{DRT}\\
\hline
EAO&0.286&0.290&0.291&0.293&0.295&0.311&0.321&0.325&0.331&\color{blue}{\bf 0.374}&\color{red}{\bf 0.442}\\
R&0.368&0.350&0.252&0.345&0.378&0.233&0.291&0.268&0.238&\color{blue}{\bf 0.200}&\color{red}{\bf 0.140}\\
A&0.557&0.496&0.465&0.541&0.544&0.490&\color{red}{\bf 0.577}&0.554&0.539&0.551&\color{blue}{\bf 0.569}\\
\hline
\label{tab:vot}
\end{tabular}
\end{center}
\end{table*}

\section{Experimental Results}
\label{sec:exp}
We demonstrate the effectiveness of the proposed tracker on the OTB-2013~\cite{WuLimYang13},
OTB-2015~\cite{wu2015object} and VOT-2016~\cite{VOT2016}
benchmark datasets.
Since our method jointly considers both discrimination and reliability for tracking, we denote it as
\textbf{DRT} for clarity.

\subsection{Implementation Details}
The proposed DRT method is mainly implemented in MATLAB and is partially accelerated with the Caffe toolkit~\cite{jia2014caffe}.
Similar with the ECO method, we also exploit an ensemble of deep (Conv1 from VGG-M, Conv4-3 from VGG-16~\cite{chatfield2014return}) and hand-crafted (HOG and Color Names) features  for target representation.
In our tracker, we use a relatively small learning rate $\omega$ (\ie 0.011) for first 10 frames to avoid model degradation with limited training samples, and
use a larger one (\ie 0.02) in the following tracking process.
The maximum number of training samples $T_{\rm max}$ and the number of fragments as  set as 50 and 9 repectively.
As to the
online joint learning formula, the trade-off parameter $\eta$ for the local consistency term is set as 1 by default and
$\theta_{\rm min}$ and $\theta_{\rm max}$ are set as 0.5 and 1.5 respectively.
One implementation of our tracker can be found in \url{https://github.com/cswaynecool/DRT}.

\subsection{Performance Evaluation}
\noindent \textbf{OTB-2013 Dataset.}
The OTB-2013 dataset~\cite{WuLimYang13} is one of the most widely used dataset
in visual tracking and contains 50 image sequences with various challenging factors.
Using this dataset, we compare the proposed DRT method with the 29 default trackers
in~\cite{WuLimYang13} and 9 more state-of-the-art trackers
including ECO~\cite{danelljan2016eco}, C-COT~\cite{danelljan2016beyond},
Staple~\cite{bertinetto2016staple}, CF2~\cite{ma2015hierarchical},
DeepSRDCF~\cite{danelljan2015convolutional}, SRDCFdecon~\cite{danelljan2016adaptive},
SINT~\cite{tao2016siamese}, SiamFC~\cite{bertinetto2016fully} and
MEEM~\cite{zhang2014meem}.
The one-pass evaluation (OPE) is employed to compare different trackers, based on two criteria
(center location error and bounding box overlap ratio).

Figure~\ref{fig:otb} (a) reports the precision and success plots of different trackers
based on the two criteria above, respectively.
Among all compared trackers, the proposed DRT method obtains the best performance, which achieves
the 95.3\% distance precision rate at the threshold of 20 pixels and a 72.0\% area-under-curve (AUC) score.

We note that it is very useful to evaluate the performance of trackers in various attributes.
The OTB-2013 dataset is divided into 11 attributes, each of which corresponds to a challenging factor
(\eg, illumination, deformation and scale change).
Figure~\ref{fig:attributes} illustrates the overlap success plots of the top 10 algorithms on 8 attributes.
We can see that our tracker achieves the best performance in all these attributes.
Specially, the proposed method improves the second best tracker ECO by 1.4\%, 2.5\%, 2.9\% and 2.5\%
in the attributes of deformation, background clutter, illumination variation and scale variation, respectively.
These results validate that our method is effective in handling such challenges.
When the object suffers from large deformations, parts of the target object will be not reliable.
Thus, it is crucial to conduct accurate reliability learning in dealing with this case.
Since our joint learning formula is insusceptible to the feature map response distributions, it can learn
the reliability score for each region more accurately.
Similarly, influenced by the cluttered background and abrupt illumination change, the feature
map inevitably highlights the background or unreliable regions in the image.
Existing CF-based algorithms learn large filter weights in such regions, thereby resulting in the
tracking failure.
In addition, these trackers usually assign most filter weights to the learned dominant regions
and ignore certain parts of the target object, which leads to inferior scale estimation performance.
By joint discrimination and reliability learning, the proposed DRT method is robust to numerous
challenges and therefore achieves a remarkable performance in comparison with other ones.

\noindent \textbf{OTB-2015 Dataset.}
The OTB-2015 dataset~\cite{wu2015object} is an extension of the OTB-2013 dataset, and contains
50 more video sequences. We also evaluate the performance of the proposed DRT method over all
100 videos in this dataset.
In our experiment, we compare with 29 default trackers in~\cite{wu2015object} and other 9
state-of-the-art trackers including ECO~\cite{danelljan2016eco}, C-COT~\cite{danelljan2016beyond},
Staple~\cite{bertinetto2016staple}, CF2~\cite{ma2015hierarchical}, DeepSRDCF~\cite{danelljan2015convolutional},
SRDCFDecon~\cite{danelljan2016adaptive}, LCT~\cite{ma2015long}, DSST~\cite{danelljan2014accurate} and
MEEM~\cite{zhang2014meem}.

Figure~\ref{fig:otb} (b) reports both precision and success plots of different trackers in terms of the OPE rule.
Overall, our DRT method provides the best result with a distance precision rate of 92.3\% and with an AUC
score of 69.9\%, which again achieves a substantial improvement of several outstanding trackers (e.g., ECO,
C-COT and DeepSRDCF).

\begin{figure}[h]
  \centering
  \begin{tabular}{c}
  \includegraphics[width=0.66\linewidth]{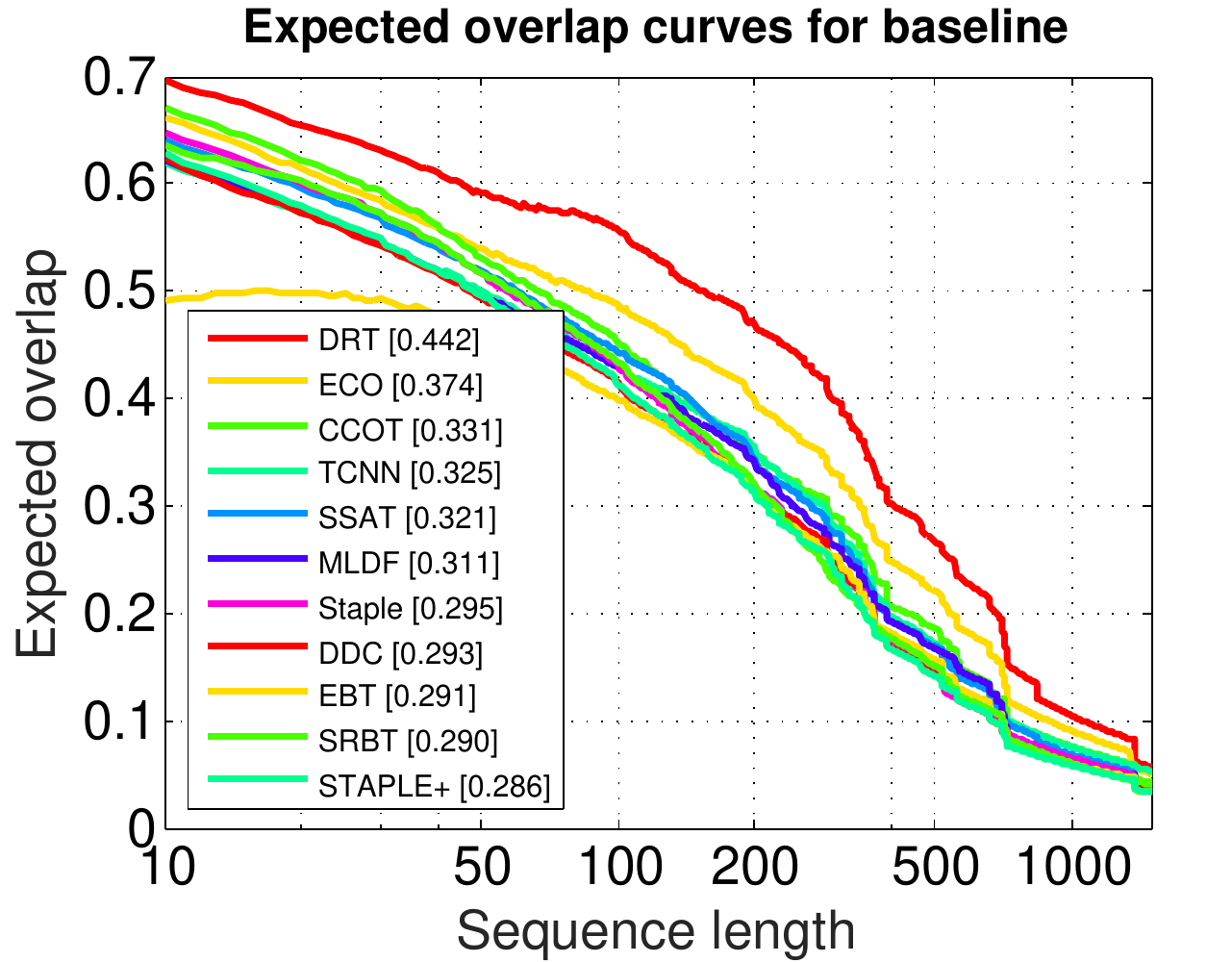}
  \end{tabular}
  \caption{Expected Average Overlap (EAO) curve for 11 state-of-the-art trackers on the VOT-2016 dataset. Our DRT tracker has much better performance than the compared trackers.}
  \label{fig:eao}
\end{figure}

\begin{figure*}[http]
\centering
\begin{tabular}{c@{}c@{}c@{}c}
\includegraphics[width=0.3\linewidth,height=42mm]{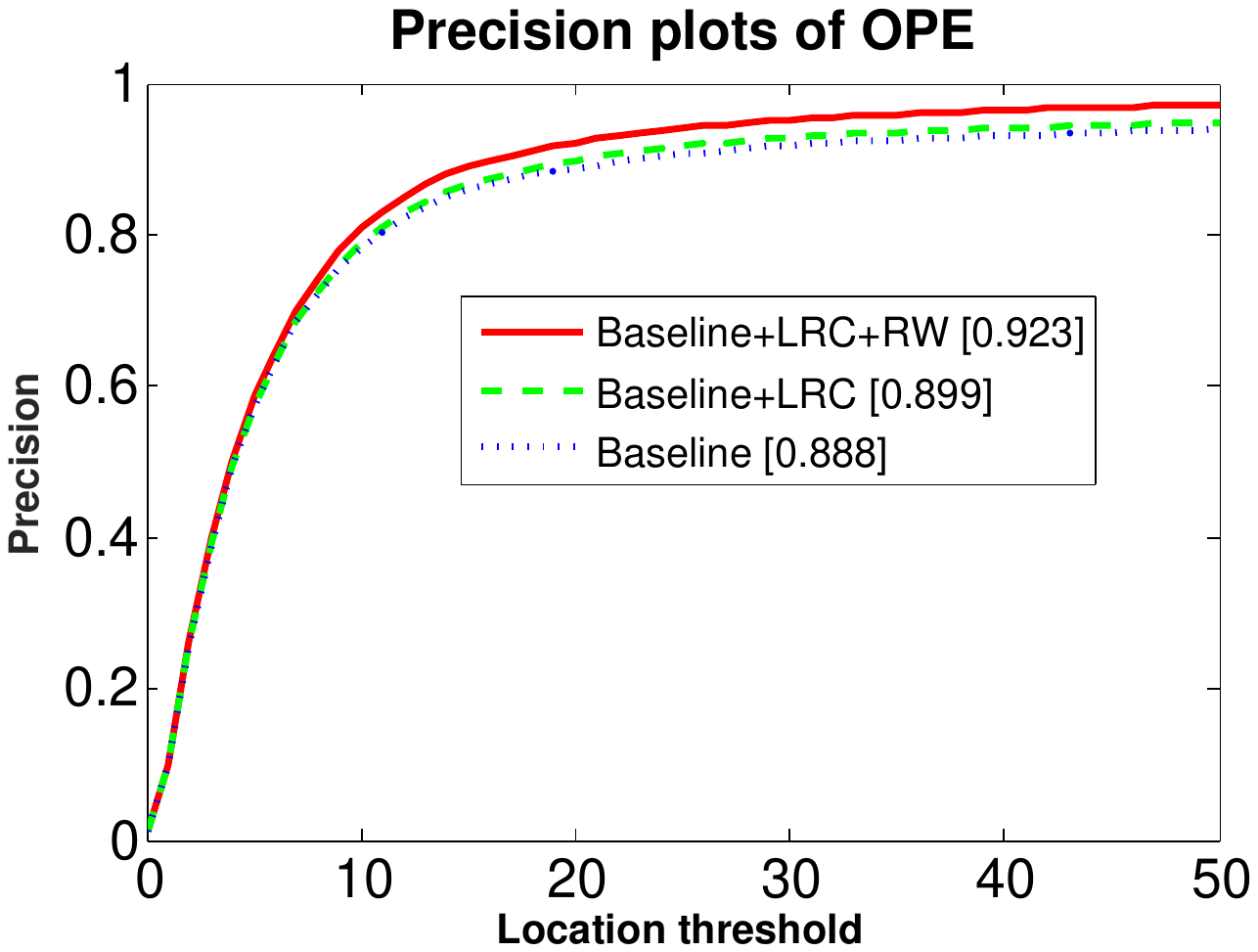}
\ &
\includegraphics[width=0.3\linewidth,height=42mm]{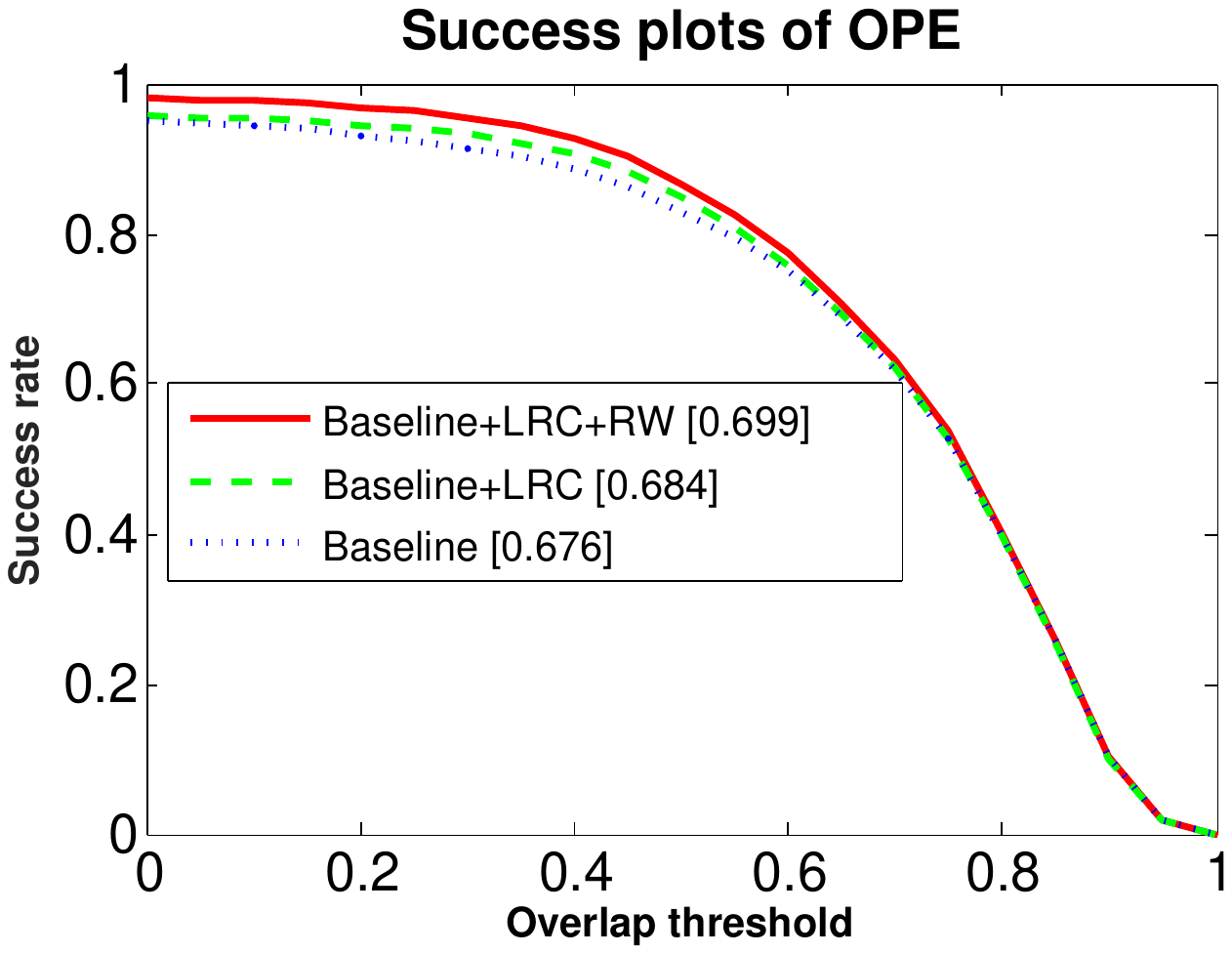}
\ &
\includegraphics[width=0.295\linewidth,height=42mm]{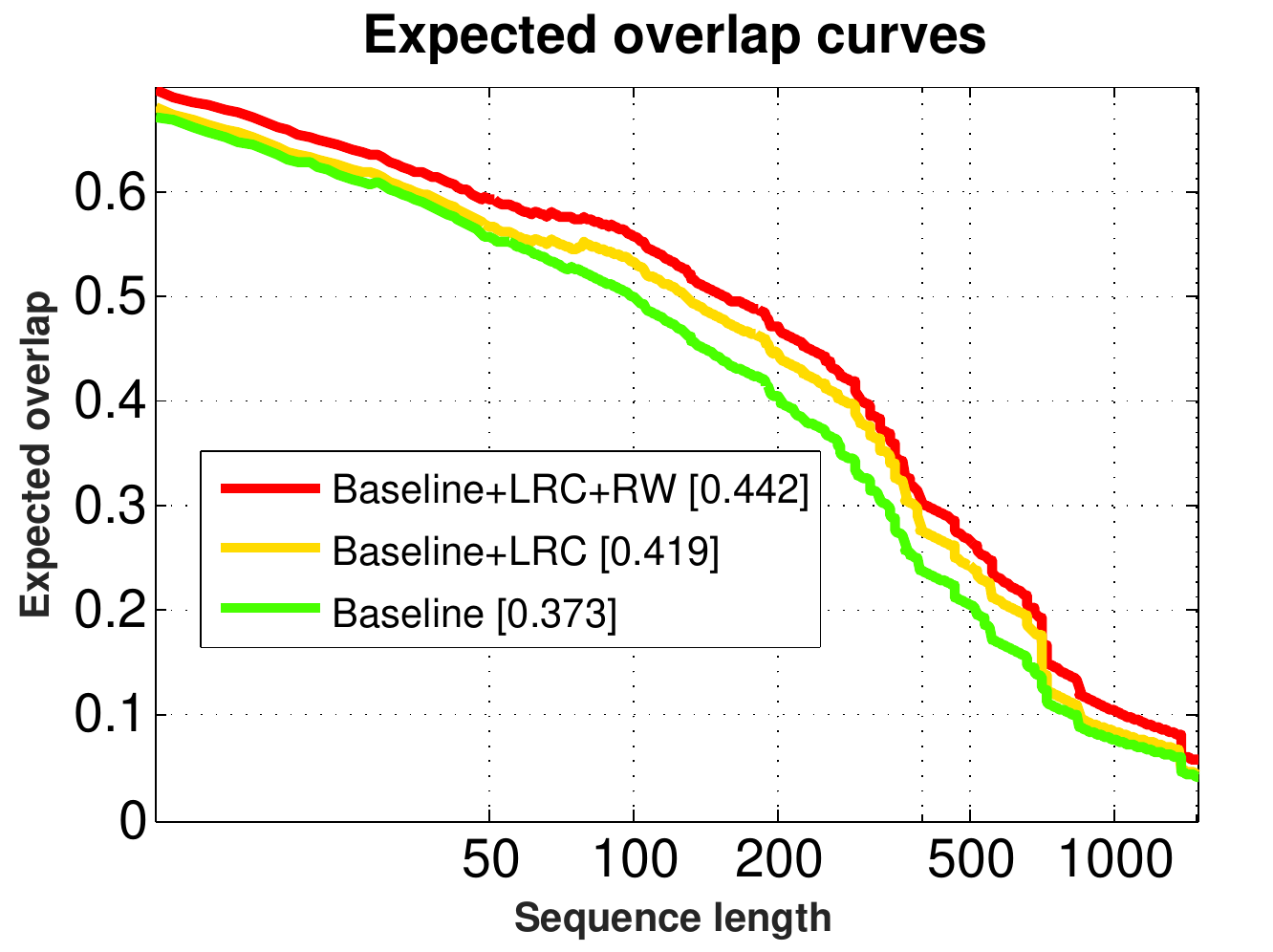}
\ \\
\PutCapablation
\end{tabular}
\caption{Performance evaluation for each component of the proposed method.}
\label{fig:ablation}
\end{figure*}

\noindent \textbf{VOT-2016 Dataset.}
The VOT-2016 dataset~\cite{VOT2016} contains 60 image sequences with 5 challenges
including camera motion, illumination change, motion change, occlusion and scale change.
Different from the OTB-2013 and OTB-2015 datasets, the VOT-2016 dataset pays much attention to
the short-term visual tracking, and thus incorporates the reset-based experiment settings.
In this work, we compare the proposed DRT method with 11 state-of-the-art trackers
including ECO~\cite{danelljan2016eco}, C-COT~\cite{danelljan2016beyond},
TCNN~\cite{nam2016modeling}, SSAT~\cite{VOT2016}, MLDF~\cite{VOT2016},
Staple~\cite{bertinetto2016staple}, DDC~\cite{VOT2016}, EBT~\cite{zhu2016beyond},
SRBT~\cite{VOT2016} and STAPLE+~\cite{bertinetto2016staple}.
The results of different tracking algorithms are reported in Table~\ref{tab:vot} (a), using the
expected average overlap (EAO), robustness raw value (R) and accuracy raw value (A) criteria.

Before our tracker, the ECO method has the best performance in the VOT-2016 dataset,
which achieves an EAO of 0.374.
Our DRT method has an EAO of $0.442$, which outperforms ECO with a relative
performance gain of $18.2\%$.
In addition, our method has the best performance in terms of robustness (\ie, fewer
failures) among all the compared methods.
Figure~\ref{fig:eao} shows the EAO curve of the compared trackers, which also demonstrates
the effectiveness of our tracker.

\subsection{Ablation Studies}
In this section, we test effectiveness for each component of the proposed joint learning formula on
both the OTB-2015 and VOT-2016 datasets. First, we use the notation ``Baseline'' to denote the
baseline method which does not exploit the local consistency constraint and the reliability map
(\ie $\beta_m=1, m\in\{1,...,M\}$). Like the conventional correlation filter, the baseline method
does not separate the discrimination and reliability information. In addition, we also use the
notation ``Baseline+LRC'' to denote the modified baseline tracker with the local response consistency
constraint. The ``Baseline+LRC'' method focuses on learning the discrimination information while
ignoring the reliability information of the target.
The abbreviation ``RW" stands for reliability weight map and ``Baseline+LRC+RW' denotes the proposed
joint learning method.
In Figure~\ref{fig:ablation}, we show that the proposed joint learning formula improves the
baseline method by 3.5\% and 2.3\% on the OTB-2015 dataset in terms of the distance precision
rate and the AUC score. In addition, the joint learning formula also improves the baseline method
by 6.9\% in EAO on the VOT-2016 dataset. By comparing our method with ``Baseline+LRC'',
we show the effectiveness of the reliability learning process. Considering the reliability learning,
 our tracker improves the ``Baseline+LRC'' method by 1.5\% in terms of AUC score on the OTB-2015
 dataset, and our tracker also improves it by 2.3\% in terms of EAO on the VOT-2016 dataset.

\vspace{-1mm}
\subsection{Failure cases} 
\begin{figure}[h]
\centering
\begin{tabular}{c@{}c@{}c}

\includegraphics[width=0.28\linewidth]{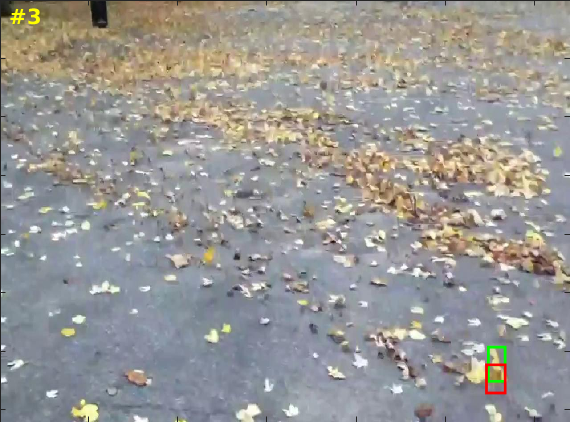}
\ &
\includegraphics[width=0.28\linewidth]{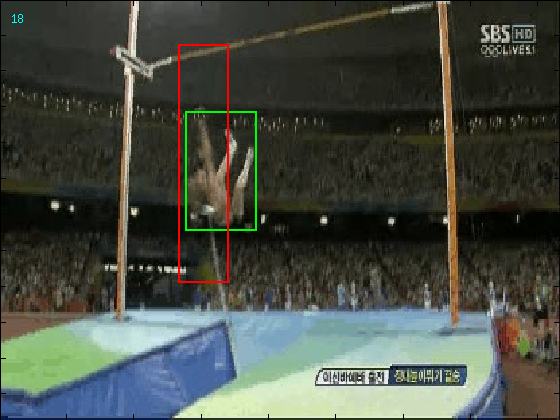}
\ &
\includegraphics[width=0.28\linewidth]{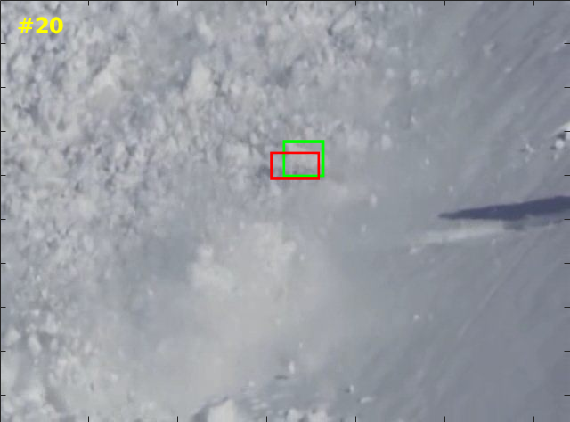}
\ \\
\includegraphics[width=0.28\linewidth]{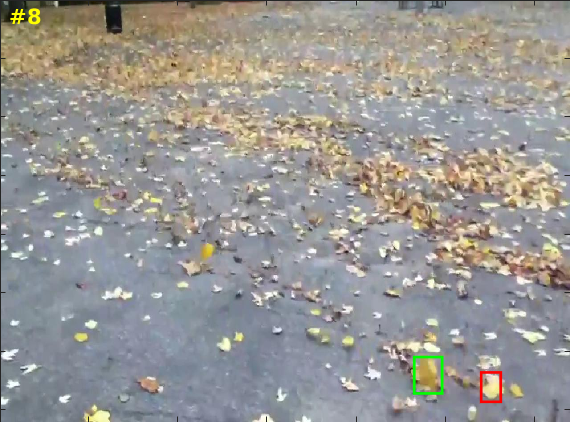}
\ &
\includegraphics[width=0.28\linewidth]{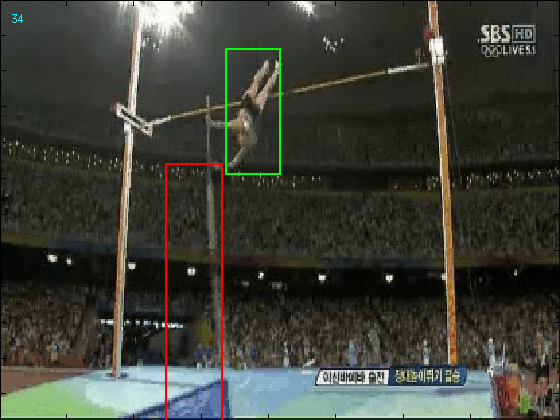}
\ &
\includegraphics[width=0.28\linewidth]{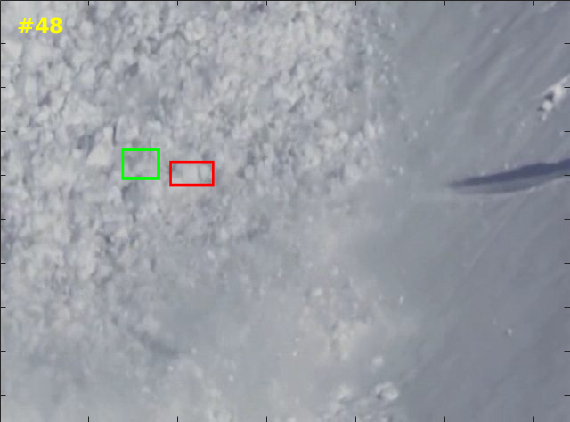}
\ \\
\end{tabular}
\caption{Failure cases of the proposed method, where we use red and green bounding boxes to denote our results and ground-truths.}
\label{fig:failure_cases}
\vspace{-2mm}
\end{figure}

We show some failure cases of the proposed tracker in Figure~\ref{fig:failure_cases}.
In the first and third columns, the cluttered background regions contain
numerous distractors, which causes the proposed method to drift off the targets.
In the second column, the proposed method does not track the target object well as it
undergoes large deformations and rotations in a short span of time. These tracking failures can
be partially addressed when the information of the optical flow is considered, which will be
the focus of our future work.

\section{Conclusion}
In this paper, we clearly consider the discrimination and reliability information in the correlation
filter (CF) formula and rewrite the filter weight as the element-wise product of a base filter and a reliability weight map.
First, we introduce a local response consistency constraint for the base filter, which constrains that
each sub-region of the target has similar importance. By this means, the reliability information is
separated from the base filter.
In addition, we consider the reliability information in the filter, which is jointly learned with the
base filter. Compared to the existing CF-based methods, our tracker is insusceptible to the non-uniform
distributions of the feature map, and can better suppress the background regions.
The joint learning of the base filter and reliability term can be preformed by solving the proposed
optimization problem and being speeded up in the Fourier domain.
Finally, we evaluate our DRT method on the OTB-2013, OTB-2015 and VOT-2016 datasets.
Extensive experiments demonstrate that the proposed tracker outperforms the state-of-the-art algorithms
over all three benchmarks.

\textbf{Acknowledgment.} This paper is partially supported by the Natural Science
Foundation of China \#61502070, \#61725202, \#61472060. Chong Sun and Ming-Hsuan Yang
are also supported in part by NSF CAREER (No. 1149783), gifts from Adobe, Toyota, Panasonic, Samsung, NEC, Verisk, and Nvidia.

{\small
\bibliographystyle{ieee}

\begin{thebibliography}{10}\itemsep=-1pt

\bibitem{bertinetto2016staple}
L.~Bertinetto, J.~Valmadre, S.~Golodetz, O.~Miksik, and P.~H. Torr.
\newblock Staple: Complementary learners for real-time tracking.
\newblock In {\em CVPR}, 2016.

\bibitem{bertinetto2016fully}
L.~Bertinetto, J.~Valmadre, J.~F. Henriques, A.~Vedaldi, and P.~H. Torr.
\newblock Fully-convolutional siamese networks for object tracking.
\newblock In {\em ECCV}, 2016.

\bibitem{bolme2010visual}
D.~S. Bolme, J.~R. Beveridge, B.~A. Draper, and Y.~M. Lui.
\newblock Visual object tracking using adaptive correlation filters.
\newblock In {\em CVPR}, 2010.

\bibitem{chatfield2014return}
K.~Chatfield, K.~Simonyan, A.~Vedaldi, and A.~Zisserman.
\newblock Return of the devil in the details: Delving deep into convolutional
  nets.
\newblock In {\em BMVC}, 2014.

\bibitem{danelljan2016eco}
M.~Danelljan, G.~Bhat, F.~S. Khan, and M.~Felsberg.
\newblock Eco: Efficient convolution operators for tracking.
\newblock In {\em CVPR}, 2017.

\bibitem{danelljan2014accurate}
M.~Danelljan, G.~H{\"a}ger, F.~Khan, and M.~Felsberg.
\newblock Accurate scale estimation for robust visual tracking.
\newblock In {\em BMVC}, 2014.

\bibitem{danelljan2015convolutional}
M.~Danelljan, G.~Hager, F.~Shahbaz~Khan, and M.~Felsberg.
\newblock Convolutional features for correlation filter based visual tracking.
\newblock In {\em ICCV Workshops}, 2015.

\bibitem{danelljan2015learning}
M.~Danelljan, G.~Hager, F.~Shahbaz~Khan, and M.~Felsberg.
\newblock Learning spatially regularized correlation filters for visual
  tracking.
\newblock In {\em ICCV}, 2015.

\bibitem{danelljan2016adaptive}
M.~Danelljan, G.~Hager, F.~Shahbaz~Khan, and M.~Felsberg.
\newblock Adaptive decontamination of the training set: A unified formulation
  for discriminative visual tracking.
\newblock In {\em CVPR}, 2016.

\bibitem{danelljan2016beyond}
M.~Danelljan, A.~Robinson, F.~S. Khan, and M.~Felsberg.
\newblock Beyond correlation filters: Learning continuous convolution operators
  for visual tracking.
\newblock In {\em ECCV}, 2016.

\bibitem{henriques2012exploiting}
J.~F. Henriques, R.~Caseiro, P.~Martins, and J.~Batista.
\newblock Exploiting the circulant structure of tracking-by-detection with
  kernels.
\newblock In {\em ECCV}, 2012.

\bibitem{henriques2015high}
J.~F. Henriques, R.~Caseiro, P.~Martins, and J.~Batista.
\newblock High-speed tracking with kernelized correlation filters.
\newblock {\em IEEE Transactions on Pattern Analysis and Machine Intelligence},
  37(3):583--596, 2015.

\bibitem{jia2014caffe}
Y.~Jia, E.~Shelhamer, J.~Donahue, S.~Karayev, J.~Long, R.~Girshick,
  S.~Guadarrama, and T.~Darrell.
\newblock Caffe: Convolutional architecture for fast feature embedding.
\newblock In {\em ICMM}, 2014.

\bibitem{kiani2017learning}
H.~Kiani~Galoogahi, A.~Fagg, and S.~Lucey.
\newblock Learning background-aware correlation filters for visual tracking.
\newblock In {\em CVPR}, 2017.

\bibitem{VOT2016}
M.~Kristan, J.~Matas, A.~Leonardis, M.~Felsberg, L.~Cehovin, G.~Fernandez,
  T.~Vojir, G.~Hager, G.~Nebehay, and R.~Pflugfelder.
\newblock The visual object tracking vot2016 challenge results.
\newblock In {\em ECCV Workshops}, 2016.

\bibitem{LiWWL18}
P.~Li, D.~Wang, L.~Wang, and H.~Lu.
\newblock Deep visual tracking: Review and experimental comparison.
\newblock {\em Pattern Recognition}, 76:323--338, 2018.

\bibitem{li2015reliable}
Y.~Li, J.~Zhu, and S.~C. Hoi.
\newblock Reliable patch trackers: Robust visual tracking by exploiting
  reliable patches.
\newblock In {\em CVPR}, 2015.

\bibitem{liu2016structural}
S.~Liu, T.~Zhang, X.~Cao, and C.~Xu.
\newblock Structural correlation filter for robust visual tracking.
\newblock In {\em CVPR}, 2016.

\bibitem{liu2015reliable}
T.~Liu, G.~Wang, and Q.~Yang.
\newblock Real-time part-based visual tracking via adaptive correlation
  filters.
\newblock In {\em CVPR}, 2015.

\bibitem{lukezic2017discriminative}
A.~Luke{\v{z}}i{\v{c}}, T.~Voj{\'\i}{\v{r}}, L.~{\v{C}}ehovin, J.~Matas, and
  M.~Kristan.
\newblock Discriminative correlation filter with channel and spatial
  reliability.
\newblock In {\em CVPR}, 2017.

\bibitem{ma2015hierarchical}
C.~Ma, J.-B. Huang, X.~Yang, and M.-H. Yang.
\newblock Hierarchical convolutional features for visual tracking.
\newblock In {\em ICCV}, 2015.

\bibitem{ma2015long}
C.~Ma, X.~Yang, C.~Zhang, and M.-H. Yang.
\newblock Long-term correlation tracking.
\newblock In {\em CVPR}, 2015.

\bibitem{nam2016modeling}
H.~Nam, M.~Baek, and B.~Han.
\newblock Modeling and propagating cnns in a tree structure for visual
  tracking.
\newblock {\em arXiv preprint arXiv:1608.07242}, 2016.

\bibitem{NoceWrig06}
J.~Nocedal and S.~J. Wright.
\newblock {\em Numerical Optimization}.
\newblock Springer, New York, USA, 2006.

\bibitem{salsc2018}
Y.~Qi, L.~Qin, J.~Zhang, S.~Zhang, Q.~Huang, and M.-H. Yang.
\newblock Structure-aware local sparse coding for visual tracking.
\newblock {\em IEEE Transactions on Image Processing}, PP(99):1--1, 2018.

\bibitem{qi2016hedged}
Y.~Qi, S.~Zhang, L.~Qin, H.~Yao, Q.~Huang, J.~Lim, and M.-H. Yang.
\newblock Hedged deep tracking.
\newblock In {\em CVPR}, 2016.

\bibitem{tao2016siamese}
R.~Tao, E.~Gavves, and A.~W. Smeulders.
\newblock Siamese instance search for tracking.
\newblock In {\em CVPR}, 2016.

\bibitem{wang2015visual}
L.~Wang, W.~Ouyang, X.~Wang, and H.~Lu.
\newblock Visual tracking with fully convolutional networks.
\newblock In {\em ICCV}, 2015.

\bibitem{wang2016stct}
L.~Wang, W.~Ouyang, X.~Wang, and H.~Lu.
\newblock Stct: Sequentially training convolutional networks for visual
  tracking.
\newblock In {\em CVPR}, 2016.

\bibitem{WuLimYang13}
Y.~Wu, J.~Lim, and M.-H. Yang.
\newblock Online object tracking: A benchmark.
\newblock In {\em CVPR}, 2013.

\bibitem{wu2015object}
Y.~Wu, J.~Lim, and M.-H. Yang.
\newblock Object tracking benchmark.
\newblock {\em IEEE Transactions on Pattern Analysis and Machine Intelligence},
  37(9):1834--1848, 2015.

\bibitem{zhang2014meem}
J.~Zhang, S.~Ma, and S.~Sclaroff.
\newblock Meem: robust tracking via multiple experts using entropy
  minimization.
\newblock In {\em ECCV}, 2014.

\bibitem{zhu2016beyond}
G.~Zhu, F.~Porikli, and H.~Li.
\newblock Beyond local search: Tracking objects everywhere with
  instance-specific proposals.
\newblock In {\em CVPR}, 2016.

\end{thebibliography}

}

\end{document}